\documentclass[11pt]{article}
\usepackage[preprint]{acl}
\usepackage{times}
\usepackage{latexsym}
\usepackage[T1]{fontenc}
\usepackage[utf8]{inputenc}
\usepackage{microtype}
\usepackage{inconsolata}
\usepackage{graphicx}
\usepackage{fontawesome5}
\usepackage{subcaption}
\usepackage{booktabs}
\usepackage{amssymb}
\usepackage{amsfonts}
\usepackage{mathtools}
\usepackage{amsmath}
\usepackage{multirow}
\usepackage{xspace}
\usepackage{xcolor}
\usepackage{enumitem}
\usepackage{tikz}
\usetikzlibrary{positioning,calc,arrows.meta}
\usepackage{algorithm}
\usepackage{algorithmic}

\newcommand{\method}{\textsc{DERL}\xspace}
\newcommand{\methodbase}{\mbox{\textsc{DERL}-base}\xspace}
\newcommand{\methodfull}{\mbox{\textsc{DERL}-full}\xspace}
\newcommand{\dev}{\mathcal{D}_{\mathrm{val}}}
\newcommand{\din}{\mathcal{D}_{\mathrm{in}}}
\newcommand{\dtrain}{\mathcal{D}_{\mathrm{train}}}
\newcommand{\dtest}{\mathcal{D}_{\mathrm{test}}}
\newcommand{\Perf}{\mathrm{Perf}}
\newcommand{\ins}{\textit{ins}}

\title{Dual-Evolving Reinforcement Learning for Automated Reward Design}

\author{
  \textbf{Sitao Cheng\textsuperscript{1*}},
  \textbf{Tianle Li\textsuperscript{2*}},
  \textbf{Xuhan Huang\textsuperscript{3*}},
  \textbf{Xunjian Yin\textsuperscript{4}},
  \textbf{Difan Zou\textsuperscript{2}}
\\
  \textsuperscript{1}University of Waterloo,
  \textsuperscript{2}The University of Hong Kong,
\\
  \textsuperscript{3}The Chinese University of Hong Kong, Shenzhen,
  \textsuperscript{4}Duke University
\\
  \small{ 
  \href{mailto:sitao.cheng@uwaterloo.ca}{sitao.cheng@uwaterloo.ca},
  \href{mailto:tianleli@connect.hku.hk}{tianleli@connect.hku.hk},
  \href{mailto:xuhanhuang@link.cuhk.edu.cn}{xuhanhuang@link.cuhk.edu.cn}}
\\
  \small{\href{https://sitaocheng.github.io/derl/}{\faGlobe~Project Page} \quad
  \href{https://github.com/sitaocheng/DERL}{\faGithub~Code} \quad
  \href{https://huggingface.co/DifferentiableEvolutionaryRL}{\raisebox{-0.15em}{\includegraphics[height=0.95em]{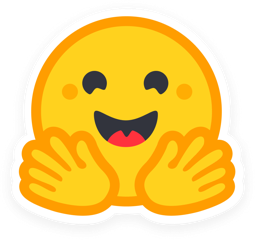}}~Models}}
}

\begin{document}
\maketitle
\begingroup\renewcommand\thefootnote{*}\footnotetext{Equal contribution.}\endgroup

\begin{abstract}
Crafting effective reward signals remains a central challenge in reinforcement learning (RL) for complex reasoning.
Existing methods for automated reward optimization search with a fixed proposer which never learns from its costly trials.
We present \textbf{Dual-Evolving Reinforcement Learning (\method)}, a dual-training framework for autonomous discovery of reward structures.
An outer-loop \emph{proposer} composes a reward from task primitives, and an inner policy trains with this reward.
The proposer itself learns by policy gradient from the validation performance of the inner policy. Thus, \method utilizes past-trial feedback to shape its reward-proposal distribution, an empirical meta-gradient of task success.
We validate \method on embodied (ALFWorld), scientific (ScienceWorld), and mathematical (GSM8K, MATH) reasoning.
On multi-turn agent benchmarks, \method outperforms strong baselines, significantly on out-of-distribution tasks. In mathematics, \method improves on the outcome reward where directly added primitives hurt.
Further analyses under matched compute show that our proposer effectively captures the meta-gradient to discover useful reward signals. 
\end{abstract}

\section{Introduction}
\label{sec:intro}

When human experts tune a learning system, they rely on an empirical \emph{meta-gradient} accumulated over past trials: the awareness of how a given change to the reward design will yield a specific shift in the behavior of the policy~\citep{knox2009interactively}.
This expertise is crucial in reinforcement learning (RL), where agent behavior depends on the reward signal: the lens through which an agent interprets environment and learns preferred behaviors~\citep{schulman2017proximal,deepseekai2025deepseekr1incentivizingreasoningcapability}.
However, crafting reward signal remains a persistent bottleneck.
On long-horizon tasks, such as embodied or scientific agent environments, a binary outcome reward is too sparse to drive learning~\citep{zhang2025rlvmr,feng2025group}, while hand-designed dense rewards invite reward hacking~\citep{amodei2016concrete,skalse2022defining,yan2025reformreducinghuman}. Learned reward models mitigate both issues~\citep{ouyang2022training} but need costly human annotation.

\begin{figure*}[t]
    \centering
    \begin{tikzpicture}[x=1pt, y=1pt, inner sep=0pt, outer sep=0pt,
        lab/.style={font=\sffamily\fontsize{6.5}{7.6}\selectfont, align=center, anchor=north},
        hdr/.style={font=\sffamily\bfseries\fontsize{7}{8}\selectfont, anchor=base},
        flow/.style={line width=1.3pt, -{Triangle[length=3.8pt, width=4.6pt]}},
        loop/.style={line width=0.9pt, -{Stealth[length=3pt, width=3.2pt]}}]
      \def\cA{0} \def\cB{69} \def\cC{142}
      \def\yTop{0} \def\yGear{-36} \def\yPol{-53}
      \node[hdr, font=\sffamily\bfseries\fontsize{8}{9}\selectfont] at (74, 27) {Reward Design Framework};
      \node[hdr] at (\cA, 15) {Traditional};
      \node[hdr] at (\cB, 15) {Static Proposer};
      \node[hdr] at (150, 15) {Dual Evolve (DERL)};
      \foreach \x in {36.5, 108.5} \draw[densely dotted, gray!70, line width=0.45pt] (\x, 12) -- (\x, -71);
      \node (top1) at (\cA, \yTop) {\includegraphics[height=20pt]{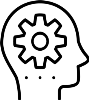}};
      \node (top2) at (\cB, \yTop) {\includegraphics[height=20pt]{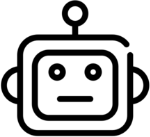}};
      \node (top3) at (\cC, \yTop) {\includegraphics[height=20pt]{images/fig1/agent_robot.png}};
      \node[anchor=north west] at ($(top2.north west)+(-4.5pt, 1pt)$) {\includegraphics[height=10pt]{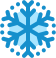}};
      \node[anchor=north west] at ($(top3.north west)+(-3pt, 1pt)$) {\includegraphics[height=9pt]{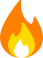}};
      \draw[loop] ($(top3.north east)+(-2.5pt, -1.5pt)$) arc[start angle=170, end angle=-80, radius=4.6pt];
      \node[lab, anchor=center] at (\cA, -20.5) {Heuristic Reward\\or Reward Model};
      \node[lab, anchor=center] at (\cB, -20.5) {Prompted Agent};
      \node[lab, anchor=center] at (\cC, -20.5) {Proposer Evolve};
      \foreach \c in {\cA, \cB, \cC} {
        \node at (\c-8, \yGear) {\includegraphics[height=9.5pt]{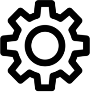}};
        \draw[flow] (\c+3, \yGear+5.5) -- (\c+3, \yGear-7.5);
      }
      \foreach \c/\n in {\cA/pol1, \cB/pol2, \cC/pol3} {
        \node (\n) at (\c, \yPol) {\includegraphics[height=16pt]{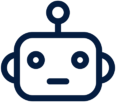}};
        \draw[loop] ($(\n.north east)+(-2pt, -2pt)$) arc[start angle=170, end angle=-80, radius=4.2pt];
        \node[lab] at (\c, \yPol-10.5) {Policy Model Evolve};
      }
      \foreach \c/\t/\f in {\cB/top2/101, \cC/top3/174}
        \draw[flow] (\c+11, \yPol-2.5) -- (\f, \yPol-2.5) -- (\f, \yTop-2) -- ($(\t.east)+(2pt, -2pt)$);
      \draw[line width=0.5pt, -{Latex[length=3.5pt, width=3pt]}] (-30, -76.5) -- (178, -76.5)
        node[midway, below=1.6pt, font=\sffamily\fontsize{6}{7}\selectfont] {Decreasing Human Labor, Increasing Automation, Increasing Scalability};
      \path let \p1 = (current bounding box.south west), \p2 = (current bounding box.north east) in
        \pgfextra{\pgfmathsetlengthmacro{\chartH}{\y2-\y1}}
        node[anchor=south west, inner sep=0pt] at (\x2+14pt, \y1) {\includegraphics[height=\chartH]{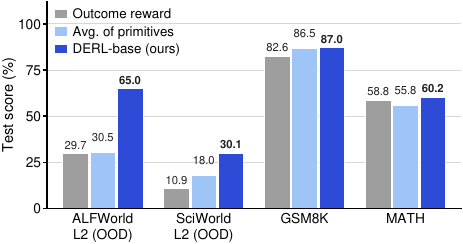}};
    \end{tikzpicture}
    \caption{\textbf{Left}: Three reward design frameworks, from most to least human labor. Early work adopts heuristics or a trained reward model. A static proposer generates and refines rewards with a frozen LLM. In \method, the proposer is itself optimized by policy gradient on the inner policy's validation performance, a dual evolution of both the policy and the reward proposer. \textbf{Right}: Test score (\%) with different rewards.}
    \label{fig:overview}
\end{figure*}

To avoid manual engineering, recent work pivots to automated reward design.
Early studies use genetic programming over primitives or gradient ascent on a designer's objective~\citep{singh2009where,niekum2010genetic,zheng2018learning}. Others ask an LLM to write and refine reward code or rubrics from execution feedback~\citep{Ma2024eureka,novikov2025alphaevolve,zhang2025darwin} or evolve training setups in a population~\citep{jaderberg2017population}. 
In these approaches, the proposer, a mutation operator or a frozen LLM, never changes: each trial costs a full RL run, yet its outcome never improves the proposer.
The accumulated feedback from reward modifications to policy performance---the empirical meta-gradient---is never retained to improve future proposals.

In this paper, we study whether a language model can learn this empirical meta-gradient from trial results. We propose \textbf{Dual-Evolving Reinforcement Learning (\method)},
a dual-training framework for autonomous discovery of reward structures (Figure~\ref{fig:overview}).
An inner-loop trains a policy with a proposed reward. An outer-loop trains the proposer, a model that generates the reward function, with the validation performance of the inner policy as its reward.
The two learners co-evolve, hence \emph{dual}.
The outer update is a score-function policy-gradient estimate~\citep{williams1992simple} that uses only sampled rewards and their scalar scores and never differentiates through inner training. The proposer thus accumulates an empirical meta-gradient: a proposal distribution shaped by earlier trials.

To facilitate dual evolution, we design a tractable search space. Rather than arbitrary code, the proposer learns to compose a set of \emph{primitives} with arithmetic operators and weights, such as executable checks, outcome signal, format validity, or the average step score over a phase of a trajectory.

We experiment on embodied (ALFWorld), scientific (ScienceWorld), and mathematical (GSM8K, MATH) benchmarks. \method outperforms strong baselines on both agent benchmarks at all three generalization levels, with significant margins on out-of-distribution tasks. In mathematics, \method improves on the outcome reward where directly added primitives hurt.
Under matched compute and the same reward space, our learned proposer outperforms static-LLMs and random search, which indicates that the gain is from the optimization of the proposer capturing the meta-gradient.
We summarize our contributions:
\begin{itemize}[leftmargin=*,itemsep=1pt,topsep=1pt]
    \item \method, a dual-evolving reward design framework where a trainable proposer learns to optimize reward functions from the validation performance of inner policy trained on its rewards.
    \item An instantiation of \method where a 0.5B proposer learns to compose symbolic primitives as reward signal in a tractable search space.
    \item Results on four benchmarks showing that \method outperforms strong baselines, and that the proposer learns to capture the meta-gradient.
\end{itemize}

\begin{figure*}[t]
    \centering
    \includegraphics[width=\linewidth]{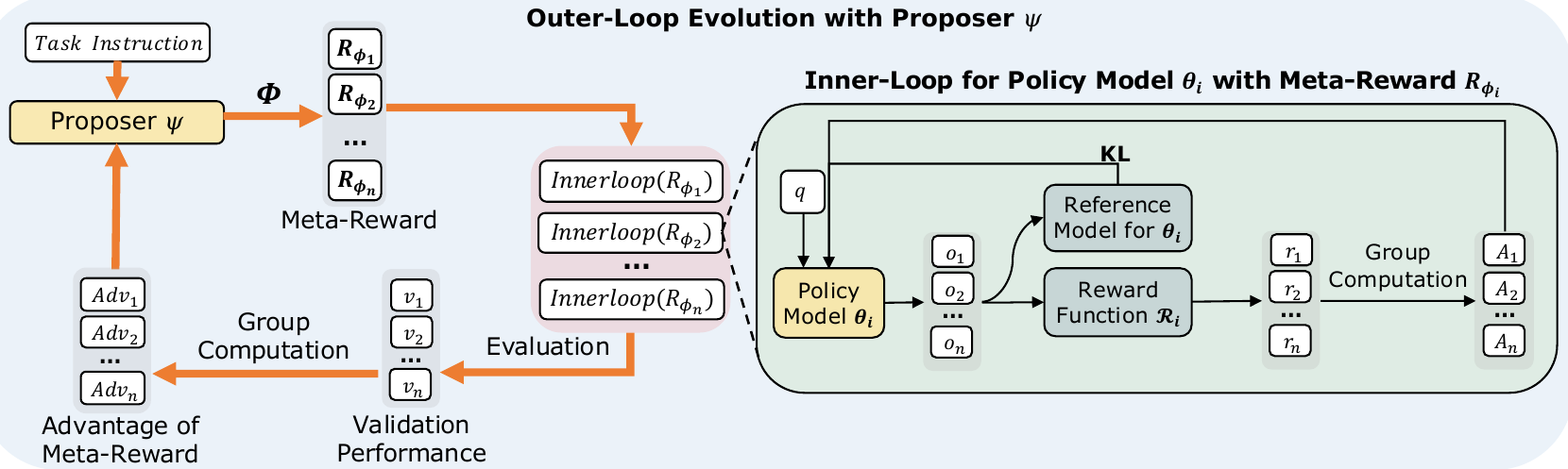}
    \caption{Our dual-evolving framework (\method). The proposer $\psi$ reads a fixed task instruction and samples $n$ reward programs $\Phi=\{\phi_1,\dots,\phi_n\}$. Each program trains one inner policy, whose validation score $v_i$ is the reward of $\phi_i$, and the group-normalized advantages update $\psi$ with GRPO. The inner-loop trains policy $\theta_i$ with GRPO under the generated reward $R_{\phi_i}$ (the index $n$ inside this block denotes the inner group size $G$ of Equation~\ref{eq:inner:adv}).}
    \label{fig:training_loop}
\end{figure*}

\section{Related Work}
\label{sec:related}

\paragraph{Reward Search.}
Outcome rewards are objective but sparse~\citep{shao2024deepseekmath,tang2025calmstormunlockingnative}.
Dense rewards come from reward models and rubric judges~\citep{ouyang2022training,lee2023rlaif,gunjal2025rubrics} or process reward models~\citep{lightman2024lets,cui2025process}, which need labels or a judge, and from step-level signals for agents~\citep{song2024trial,xiong2024watch,zhang2025rlvmr,feng2025group}.
Hand-designed reward terms need manual balancing~\citep{yu2025dapo,yan2025reformreducinghuman}.
Reward search optimizes the reward itself for the designer's objective~\citep{singh2009where}, by evolution~\citep{niekum2010genetic,faust2019evolving} or by gradients through the agent's update~\citep{zheng2018learning}.
\method searches the reward function by composing primitives on signals from a validation split.

\paragraph{Automatic Agent Evolution.}
LLM agents~\citep{qin2023toolllm,huang-etal-2024-queryagent,cheng-etal-2024-call,an-etal-2025-thread} still rely on hand-engineered prompts, workflows, and rewards~\citep{sutton2019bitter}.
Evolutionary methods search by perturbation and selection over policy weights~\citep{such2017deep,khadka2018evolution} or hyperparameters ~\citep{jaderberg2017population}.
Prompted LLMs now perturb programs~\citep{Romera24mathematical,novikov2025alphaevolve}, agents~\citep{zhang2025darwin}, and reward code or rubrics~\citep{xie2024text2reward,Ma2024eureka,shao2025dr} from training feedback.
In these methods the proposer is static, and a trial's score acts only through selection or the prompt, never entering training~\citep{gao2025survey,cheng2026xtree}.
In \method, we optimize the proposer itself by policy gradient on these scores, so its weights store the accumulated feedback of all past trials, capturing the meta-gradient.

\paragraph{Learning to Learn.}
Meta-learning improves a learner with experience~\citep{vilalta2002meta-survey,andrychowicz2016learninglearngradientdescent}.
In RL, meta-learners tune the return, update rule, auxiliary rewards, or sampling temperature~\citep{xu2018meta,oh2020discovering,agarwal2019learning,dang2026temperature}, often by differentiating through the learner's update.
Another line trains a proposer on the performance of its outputs: architectures and optimizers by RL~\citep{zoph2017neural,Bello2017Neural}, the program proposer of EvoTune~\citep{surina2025algorithm}, and the proposer--verifier pair of RL Tango~\citep{zha2025rltango}.
\method trains such a proposer for reward functions on the validation performance of inner-loop policy using the proposed reward, forming a dual-evolving framework.

\section{Dual-Evolving Reinforcement Learning}
\label{sec:method}

We formulate \method as a dual-evolving process in which both the policy and the reward proposer learn, and then describe its instantiation.

\subsection{Overview}
\label{sec:formulation}
\method replaces the fixed reward proposer with one that learns.
The outer-loop trains the proposer $\psi$ by RL to generate a configuration $\phi$, a reward program such as an arithmetic expression over primitives (\S\ref{sec:instantiation}), which specifies a reward function $R_\phi$ called \emph{Meta-Reward} (Figure~\ref{fig:training_loop}).
The inner-loop trains a policy $\pi_\theta$ under $R_\phi$. Given a task $q$, the policy produces an output $o$, a full interaction trajectory in agent tasks.
We split the original training set into an inner-training split $\din$ and a validation split $\dev$.
Algorithm~\ref{alg:derl} in Appendix~\ref{app:algorithm} lists the full procedure.

\paragraph{Inner-loop (policy evolution).}
Given $\phi$, the inner policy maximizes the Meta-Reward $R_\phi$ on $\din$:
\begin{equation}
\label{eq:inner:objective}
\mathcal{J}^{\mathrm{in}}_{\phi}(\theta) = \mathbb{E}_{q \sim \din,\, o \sim \pi_{\theta}(\cdot \mid q)} \big[R_{\phi}(o)\big].
\end{equation}

\paragraph{Outer-loop (proposer evolution).}
The proposer policy $\pi_\psi(\phi \mid \ins)$ reads a task instruction $\ins$ and writes a configuration $\phi$.
The inner policy trained with $R_\phi$ is scored on $\dev$ by the task metric $\Perf$ (\textit{e.g.,} success rate, accuracy), and the proposer policy maximizes the expected score:
\begin{equation}
\label{eq:outer:objective}
\mathcal{J}^{\mathrm{out}}(\psi) = \mathbb{E}_{\phi \sim \pi_{\psi}} \big[\Perf(\theta^{\star}(\phi); \dev)\big],
\end{equation}
with $\theta^{\star}(\phi) = \arg\max_{\theta} \mathcal{J}^{\mathrm{in}}_{\phi}(\theta)$, approximated in practice by the policy after a fixed budget of inner training.
Because $\phi$ is discrete and its validation performance $v(\phi) = \Perf(\theta^{\star}(\phi); \dev)$ is obtained only after a full inner RL run, we apply the score-function identity (REINFORCE;~\citealp{williams1992simple}), which estimates the gradient from samples and their scalar values alone:
\begin{equation}
\label{eq:outer:gradient}
\nabla_{\psi}\mathcal{J}^{\mathrm{out}} = \mathbb{E}_{\phi \sim \pi_{\psi}}\big[v(\phi) \nabla_{\psi} \log \pi_{\psi}(\phi \mid \ins)\big].
\end{equation}

\paragraph{What the proposer learns.}
In \method, the discrete search over the reward configuration $\phi$ becomes the policy-gradient optimization of continuous proposer parameters $\psi$. The inner training and evaluation form the environment of the proposer (Appendix~\ref{app:gradient_prop}).
Each update shifts $\pi_\psi$ toward the configurations whose inner policies outperform those of the others sampled in the same outer step.
Through dual evolution, $\pi_\psi$ thus encodes the empirical meta-gradient: which reward configurations lead to higher validation performance.

\subsection{Reward Structure Design}
\label{sec:instantiation}

A central aspect of the dual evolution is a tractable search space for the proposer output $\phi$, which parameterizes the reward function of the inner policy.
Free-form function generation makes the outer action space unbounded and leaves some samples non-executable.
In \method, the proposer instead composes a set of $k$ \emph{primitives} $g_1,\dots,g_k$.
Each primitive is an executable function that scores one aspect of the output $o$ given a context $\mathcal{C}$, such as the task $q$ and its ground truth $a^*$.
Example primitives include binary outcome correctness (\textit{i.e.,} comparing $o$ to $a^*$), adherence to formatting rules, or process heuristics such as step counts.
The proposer generates only the \textit{structure and weights} of the reward function as $\phi$, an arithmetic expression over the primitives.
With $\mathrm{Func}$ applying the weights and operators of $\phi$ (\textit{e.g.,} $+$, $-$, $\times$, $/$) to the primitive scores, the Meta-Reward is instantiated by
\begin{equation}
\label{reward_fun}
R_{\phi}(o) = \mathrm{Func}\big(g_1(o, \mathcal{C}), \dots, g_k(o, \mathcal{C}); \phi\big).
\end{equation}
With the primitives of each domain (\S\ref{sec:primitives}), the outcome, outcome-plus-format, and average baselines are points in the search space.
This ensures an expressive search space while focusing the proposer on structural reasoning rather than raw text parsing. By decoupling the definition of atomic signals from their utilization, \method generalizes to new tasks with a few lines of primitive code. The dual-evolving framework learns to optimize the composition of components, discarding detrimental primitives without manual checks (\S\ref{sec:sensitivity}). Appendix~\ref{app:primitive_design} gives more details and examples.

\subsection{Training of \method}
\label{sec:training}
\paragraph{Inner-loop training.}
We train the inner policy (green block in Figure~\ref{fig:training_loop}) with Group Relative Policy Optimization (GRPO;~\citealp{shao2024deepseekmath}).
For each task $q \sim \din$, we sample $G$ outputs $\{o_i\}_{i=1}^{G}$ from the old policy $\pi_{\theta_{\mathrm{old}}}$ and score them with the Meta-Reward, $r_i = R_\phi(o_i)$.
The advantage of $o_i$ is its reward normalized within the group:
\begin{equation}
\label{eq:inner:adv}
A_i = \frac{r_i - \mathrm{mean}(\{r_j\}_{j=1}^{G})}{\mathrm{std}(\{r_j\}_{j=1}^{G})}.
\end{equation}
With probability ratio $\rho_i = \pi_\theta(o_i \mid q) / \pi_{\theta_{\mathrm{old}}}(o_i \mid q)$, clipping range $\epsilon$, and KL divergence $D_{\mathrm{KL}}$ between $\pi_\theta$ and a reference policy $\pi_{\mathrm{ref}}$, weighted by $\beta$, GRPO maximizes
\begin{equation}
\label{eq:grpo}
\begin{aligned}
\mathcal{J}^{\mathrm{GRPO}}_{\phi}(\theta) &= \mathbb{E}_{q,\{o_i\}} \Big[ \frac{1}{G} \sum_{i=1}^{G} \min\big(\rho_i A_i, \\
&\quad \mathrm{clip}(\rho_i, 1{\pm}\epsilon)\, A_i\big) \Big] - \beta D_{\mathrm{KL}}.
\end{aligned}
\end{equation}
The Meta-Reward is the only change from standard GRPO.
We use two initialization schemes for the inner policy.
\textbf{\methodbase} starts every inner policy from the same base model $\theta_0$ for all outer step proposals.
\textbf{\methodfull} divides the baseline training budget into outer rounds. Within each round, all candidates share a starting checkpoint (the base model in the first round) and train under the proposed rewards. The best-scoring inner policy from the previous round is used to initialize the inner policies for the current round, as in population-based training~\citep{jaderberg2017population}. \methodfull thus finds a reward function per round of training.

\begin{table*}[t]
\centering
\setlength{\tabcolsep}{10pt}
\begin{tabular}{lcccccc}
\toprule
\multirow{2}{*}{\textbf{Method}} & \multicolumn{3}{c}{\textbf{ALFWorld}} & \multicolumn{3}{c}{\textbf{ScienceWorld}}\\
\cmidrule(lr){2-4} \cmidrule(lr){5-7}
 & \textbf{L0} & \textbf{L1} & \textbf{L2 (OOD)} & \textbf{L0} & \textbf{L1} & \textbf{L2 (OOD)}\\
\midrule
GRPO w/ Outcome & 76.6 & 71.1 & 29.7 & 21.1 & 13.7 & 10.9\\
GRPO w/ Average(primitives) & 88.1 & 85.4 & 30.5 & 37.9 & 31.3 & 18.0\\
GiGPO  & 86.7 & 83.2 & 48.0 & 25.8 & 15.2 & 4.7\\
RLVMR  & 89.1 & 87.9 & 56.3 & 46.9 & 34.4 & 26.5\\
\midrule
\methodbase & \underline{91.0} & \textbf{89.1} & \underline{65.0} & \underline{47.7} & \underline{43.0} & \underline{30.1}\\
\methodfull & \textbf{91.8} & \underline{88.3} & \textbf{76.4} & \textbf{98.2} & \textbf{95.3} & \textbf{31.3}\\
\bottomrule
\end{tabular}
\vspace{-5pt}
\caption{Success rate (\%) on ALFWorld and ScienceWorld with \texttt{Qwen2.5-1.5B-Instruct} at three levels as the mean over four seeds (statistics in Appendix~\ref{app:statistics_of_agent}).
GiGPO and RLVMR results are from \citet{zhang2025rlvmr}. \textbf{Bold}: best. \underline{Underline}: second best. 
\methodbase beats all baselines at every level, significantly on OOD tasks.}
\vspace{-10pt}
\label{tab:results_on_agent}
\end{table*}

\paragraph{Outer-loop training.}
The outer-loop trains the proposer with GRPO as well (blue block in Figure~\ref{fig:training_loop}).
In each of the $E_{\mathrm{out}}$ outer steps, the proposer samples $n$ configurations $\phi_1,\dots,\phi_n$ from $\pi_\psi(\cdot \mid \ins)$, and each $\phi_i$ trains an inner policy $\theta_i$ on $\din$ by Equation~\eqref{eq:grpo}.
The $n$ inner runs execute in parallel, and each $\theta_i$ is scored on $\dev$ by task success as validation performance $v_i = \Perf(\theta_i;\dev)$.
With $f_{\theta}(q)$ the output of $\pi_\theta$ for task $q$ under deterministic decoding (a final answer in mathematics, a full trajectory in agent tasks) and $\mathrm{Succ}(\cdot) \in \{0,1\}$ the success check of the task (answer match or completion reported by the environment),
\begin{equation}
\label{eq:perf}
\Perf(\theta;\dev) = \mathbb{E}_{q \sim \dev}\, \mathrm{Succ}\big(f_{\theta}(q), q\big).
\end{equation}
Let $\bar{v}$ and $\sigma_v$ be the mean and standard deviation of $\{v_1,\dots,v_n\}$.
The group-normalized score $\hat{A}_i = (v_i - \bar{v})/\sigma_v$ is the advantage of every token of $\phi_i$ in the clipped GRPO objective with a KL penalty, which estimates Equation~\eqref{eq:outer:gradient}. Notably, the validation score $v$ only rewards the proposer and is not reported. Instead, \methodbase finds the reward function with the highest $v$ in the last outer step, while \methodfull keeps one function per round. We then train a fresh policy with the found reward on the full training set under the baseline budget, and evaluate this policy on held-out test data (\S\ref{sec:protocol}).
The search of \methodbase costs $n \times E_{\mathrm{out}}$ inner runs, whereas \methodfull splits one training budget across its rounds and costs $n$ full runs (Appendix~\ref{app:cost_ana}).

Crucially, \method establishes a closed loop that returns non-differentiable validation signals to the proposer to capture the meta-gradient (more in Appendix~\ref{app:gradient_prop}). This RL-driven feedback loop replaces manual heuristics with an autonomous optimization path, enabling the composition of dense, actionable signals from sparse outcomes to resolve scalability challenges in complex domains.

\section{Experiments}
\label{sec:experiments}


We first describe the benchmarks and baselines (\S\ref{sec:exp_setup}), the protocol (\S\ref{sec:protocol}), and the primitives (\S\ref{sec:primitives}).
We then test whether the reward found by \method trains better policies than common rewards (\S\ref{sec:results}).

\subsection{Benchmarks and Baselines}
\label{sec:exp_setup}
\paragraph{Embodied agents.}
ALFWorld~\citep{shridhar2020alfworld} is a text-based household environment with six task categories.
The official test set has a \emph{seen} split, which places the objects and receptacles of training in new positions or quantities, and an \emph{unseen} split, which uses object variants and instructions never seen in training.
Following \citet{zhang2025rlvmr}, we test three generalization levels.
\textbf{L0} (in-distribution, seen) trains on all six tasks and tests on the seen split.
\textbf{L1} (in-distribution, unseen) trains on all six and tests on the unseen split.
\textbf{L2} (out-of-distribution, unseen) trains on four and tests on two held-out tasks (detailed in Appendix~\ref{app:generalization_settings}).
We compare \method with four baselines.
\textbf{1)} {GRPO w/ Outcome} uses the binary outcome reward.
\textbf{2)} {GRPO w/ Average(primitives)} uses the uniform average of all primitives (\S\ref{sec:primitives}), a common way to combine several reward terms.
\textbf{3) }{GiGPO}~\citep{feng2025group} adds step-level credit assignment to GRPO.
\textbf{4) }{RLVMR}~\citep{zhang2025rlvmr} adds a hand-designed process reward and is the strongest published method on ALFWorld.
Learned reward models are left out, because they need extra training and the outcome is verifiable. We use Qwen2.5-1.5B-Instruct as inner policy.
\methodbase runs 16 outer steps with inner runs of 40 epochs. The final policy trains with the found reward for 100 epochs, as in RLVMR, while the other baselines train for 150.
\methodfull runs 10 rounds of 10 epochs, which matches the 100-epoch budget.

\paragraph{Scientific simulation.} 
ScienceWorld~\citep{wang2022scienceworld} is an interactive text environment with elementary-school science tasks, such as measuring a melting point or growing a plant.
We use three analogous generalization levels and the same baselines as for ALFWorld.
\methodbase runs 5 outer steps with inner runs of 80 epochs.
\methodfull runs 3 rounds of 33 epochs, which matches the 100-epoch budget.

\paragraph{Mathematical reasoning.} We train with MATH only or with MATH and GSM8K, and evaluate exact match on MATH~\citep{hendrycks2021measuring} and GSM8K~\citep{cobbe2021gsm8k}.
We compare three baseline reward functions: \textbf{1)} the binary {Outcome reward}, \textbf{2)} {Outcome\,+\,Format}, which adds the format primitive, and \textbf{3)} {Average(primitives)}, the uniform average of all primitives.
The inner policy is \texttt{Qwen2.5-3B}.
\methodbase runs 8 outer steps with inner runs of 10 epochs.
The final policy trains with the found reward for 15 epochs, like the baselines.
\methodfull runs 7 steps of 2 epochs.


\subsection{Protocol}
\label{sec:protocol}
For reward search, we randomly split the original training set 8:2 into $\din$ and $\dev$ for inner training and validation, respectively.
After search, we train a fresh policy on the full training set. \methodbase uses the reward with the highest validation score in the last step. \methodfull uses each round's selected reward in the original order and with the same per-round budgets. The final policy is evaluated on held-out test set, never used during search.
The final policy receives identical data and no more training epochs than baselines.
Agent results are the mean over four seeds, as in \citet{zhang2025rlvmr}.
Mathematics results are the mean over three seeds.
We run GRPO with eight rollouts by default.
We use \texttt{Qwen2.5-0.5B-Instruct} as the proposer.

\subsection{Primitive Design}
\label{sec:primitives}
\paragraph{Agents.}
The four primitives score the outcome and the reasoning steps.
$g_1$ checks if the task succeeds.
Following RLVMR~\citep{zhang2025rlvmr}, the policy tags each action, then a rule-based check produces a step score.
$g_2$, $g_3$, and $g_4$ are the mean step scores over the first, middle, and last third of the trajectory, so the reward function can weight early, middle, and late behavior differently.
For example, the function $g_1 + (g_2 \cdot (g_3/2)) - (g_4/3)$ gives a reward of $1.25$ to a successful six-step trajectory with step scores $[1,0,1,1,0,0]$, whose primitives are $(1, 0.5, 1, 0)$ (more examples in Appendix~\ref{app:primitive_examples}).

\paragraph{Mathematics.}
$g_1$ checks if the output answer matches the ground truth.
$g_2$ checks if the answer is enclosed in \texttt{\textbackslash boxed\{\}}. $g_3$ checks if step markers such as ``Step 1'' appear.
$g_4$ checks if the ground truth appears anywhere in the output, which rewards a correct answer in the wrong format.
These primitives are simple by design. \S\ref{sec:sensitivity} studies the robustness of varied primitive design.

\subsection{Main Results}
\label{sec:results}

\begin{table}[t]
\centering
\small
\setlength{\tabcolsep}{12pt}
\begin{tabular}{lcc}
\toprule
\textbf{Reward Function} & \textbf{GSM8K} & \textbf{MATH} \\
\midrule
\multicolumn{3}{c}{\textit{Trained on MATH}} \\
Outcome      & 82.9 & 59.1 \\
Outcome + Format & 83.9 & 56.8 \\
Average(primitives)         & 83.6 & 54.9 \\
\methodbase & 83.2 & \underline{60.5} \\
\methodfull & 84.1 & \textbf{60.9} \\
\midrule
\multicolumn{3}{c}{\textit{Trained on MATH and GSM8K}} \\
Outcome      & 82.6 & 58.8 \\
Outcome + Format & 86.4 & 55.9 \\
Average(primitives)  & 86.5 & 55.8 \\
\methodbase & \underline{87.0} & 60.2 \\
\methodfull & \textbf{87.6} & 60.2 \\
\bottomrule
\end{tabular}
\caption{Exact match (\%) on GSM8K and MATH with \texttt{Qwen2.5-3B}. \textbf{Bold}: best. \underline{Underline}: second best.}
\label{tab:results_on_math}
\end{table}

\paragraph{Agents.}
Table~\ref{tab:results_on_agent} shows the success rate on ALFWorld and ScienceWorld.
\methodbase surpasses all baselines at every level on both benchmarks.
Its margin over RLVMR, the previous state of the art, reaches 8.7\% on ALFWorld L2 and 8.6\% on ScienceWorld L1, with no more human labor than RLVMR.
\methodfull reaches 91.8\% on ALFWorld L0 and 98.2\% on ScienceWorld L0 (training curves in Appendix~\ref{app:training_dynamics_of_pop}).
The gap between Average(primitives) and the outcome reward shows that the primitives are useful.
\methodbase still gains substantially over Average(primitives), so the learned weights and structure add value.
It more than doubles the success rate of Average(primitives) on ALFWorld L2 and leads by up to 11.7\% on ScienceWorld L0 and L1.
Notably, \methodbase surpasses RLVMR, the previous state-of-the-art method, at all levels without introducing any more human labor. This shows that the proposer in \method effectively captures the reward signal that helps task success.

\paragraph{Out-of-distribution (OOD) generalization.}
On ALFWorld, Average(primitives) beats the outcome reward by up to 14.3\% at the in-distribution levels but by only 0.8\% at L2. This indicates that uniform averaging fits the training tasks without transferring to new ones.
However, \methodbase and \methodfull more than double the outcome reward on OOD tasks on both benchmarks. This shows that the policies trained with the found rewards transfer to task that the search never saw.

\begin{figure}[t]
\centering
\includegraphics[width=\linewidth]{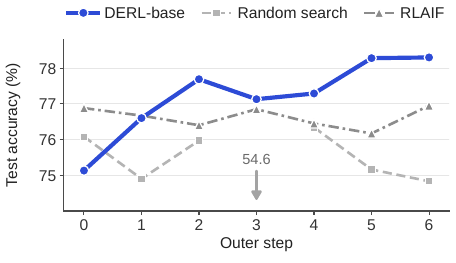}
\caption{Measn GSM8K test exact match (\%) of the inner policies (\texttt{Qwen2.5-1.5B}) trained with the proposals of each outer step. RLAIF stays flat. Random search fluctuates and ends lower. only \method improves, because its proposer learns.}
\label{fig:proposers}
\end{figure}

\begin{figure*}[t]
\centering
\includegraphics[width=\linewidth]{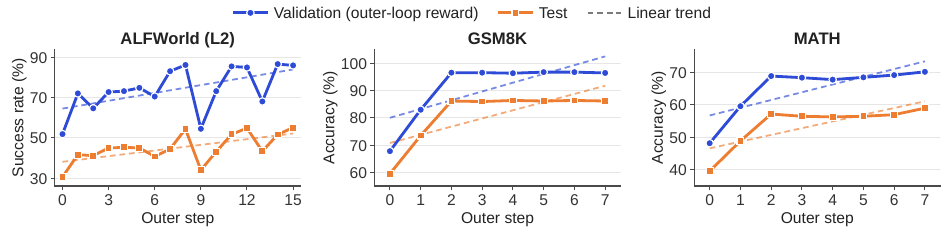}
\caption{Outer-loop dynamics on ALFWorld (L2), GSM8K, and MATH. The x-axis is the outer step. Each point is the mean over the $n$ inner policies trained from that step's proposals, scored on the validation split (blue, the outer-loop reward) and on the test set (orange, never used for training). Dashed lines are linear trends.}
\label{fig:analysis}
\end{figure*}

\paragraph{Mathematics.}
Reward search is harder in this domain, because the outcome reward is already a reliable signal. Table~\ref{tab:results_on_math} shows the performance with different reward functions.
Directly adding the format and averaged rewards raises GSM8K by up to 3.9\% but lowers MATH by up to 4.2\%. This indicates that the policy trades reasoning for formatting under uniform rewards.
However, the reward found by \method avoids this trade-off.
\methodbase exceeds the outcome reward on both benchmarks under both training conditions, and \methodfull scores higher still in three of the four cases. This shows that \method enables the navigation of the delicate trade-off between signal density and fidelity by autonomously optimizing the reward structure.

\section{Analysis}
\label{sec:analysis}

We first compare the learned proposer with static proposers (\S\ref{sec:learned_proposer}).
We then examine the outer-loop dynamics (\S\ref{sec:opt_dynamics}), the robustness to the primitives (\S\ref{sec:sensitivity}), and the structures of the rewards that the proposer writes (\S\ref{sec:structures}). Appendix \ref{app:number_of_rollouts} shows the sample efficiency of the proposer evolution.

\subsection{Does Evolution of the Proposer Matter?}
\label{sec:learned_proposer}
We study whether a static proposer can do as well with the same search budget.
We compare three proposers over the {same} reward space on GSM8K with a \texttt{Qwen2.5-1.5B} inner policy under the same settings (Appendix~\ref{app:additional_experiments}).
\textbf{Random Search} samples valid reward structures from the initial proposer and never updates it.
\textbf{RLAIF} uses GPT-4o as a prompted proposer in the style of \citet{Ma2024eureka}.
At each outer step, GPT-4o reads the reward functions and scores of the previous step and writes new ones, with no parameter update. Our \textbf{\methodbase} trains a 0.5B proposer, which is much smaller than the closed-sourced GPT-4o.



Figure~\ref{fig:proposers} shows the mean test accuracy of the inner policies trained with the proposals ($n=8$) of each outer step. The final policies trained with the best reward function of each proposer reach 79.2\% (\methodbase at last step), 77.2\% (RLAIF at last step), and 76.5\% (random search, best over all steps), against 75.8\% for outcome reward. The evolution of the proposer captures an empirical meta-gradient, learning to generate better reward structures.
The random search fluctuate and drop from 76.1\% to 74.8\%, with a collapse at step 3, while RLAIF stays between 76.2\% and 77.0\%.
Only the proposals of \methodbase improve, from 75.1\% to 78.3\%, as its proposer is updated. And updating the 0.5B proposer on these few samples costs far less than calling a proprietary model. Therefore, the gain of \method comes from optimizing the proposer on the inner policy validation performance, not from extra search budget.

\subsection{Outer-loop Training Dynamics}
\label{sec:opt_dynamics}

\S\ref{sec:learned_proposer} shows that our \method proposer truly learns to generate a better reward functions than random search. We now study \textit{whether the proposer improves its rewards steadily over outer steps}.
Figure~\ref{fig:analysis} tracks the mean validation and test performance of $n$ inner policies at each outer step on ALFWorld, GSM8K, and MATH.
We omit ScienceWorld since it converges within five outer steps.

We observe a \textit{consistent upward trend} in the mean performance of both validation and testing as the outer-loop progresses. Specifically, the trend in mathematical reasoning is more stable as the proposer recognizes that the verifiable task is mainly driven by the outcome reward. For agent tasks, outer-loop optimization involves more exploration than exploitation, ultimately showcasing robust evolution.
The concurrent rise in validation and testing performance shows that the proposer does \textit{not overfit} to the validation split that supplies its reward. Instead, it successfully approximates the meta-gradient of task success. Leveraging the validation performance, the \method proposer gradually refines the reward function, generating increasingly effective signals that guide the inner-loop agent toward higher performance. Appendix~\ref{app:outer_loop_examples} provides concrete examples of Meta-Rewards explored across outer-loop iterations.



\subsection{Robustness to Varied Primitives}
\label{sec:sensitivity}

\begin{figure}[t]
\centering
\includegraphics[width=\linewidth]{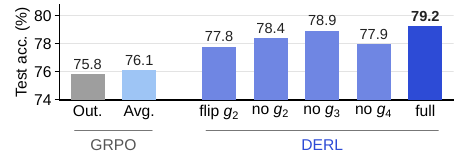}
\caption{GSM8K exact match (\%) when a primitive is removed or flipped (\texttt{Qwen2.5-1.5B} inner policy).}
\label{fig:vocab}
\end{figure}


A concern in automated reward design is the need of hand-crafted primitive signals. 
To demonstrate that \method does not rely on carefully engineered primitives and exhibits robustness to sub-optimal inputs, we evaluate with varied primitives on GSM8K. 
Relative to the outcome reward, the mathematics vocabulary adds three primitives ($g_2$ format, $g_3$ step markers, $g_4$ soft match). 
We remove each in turn and, separately, replace $g_2$ by $-g_2$ so that correct formatting is penalized (``flip $g_2$''). 

Figure~\ref{fig:vocab} shows that removing one primitive costs at most $1.3\%$, and every reduced or flipped vocabulary still beats both GRPO baselines by up to $3.1\%$. 
With the flipped primitive, the proposer learns a negative coefficient for it, which recovers the useful direction of the signal. 
This confirms that even when primitives are removed or intentionally corrupted, the outer proposer can still consistently discover and compose effective reward functions.

\subsection{Dynamics of Reward Structures}
\label{sec:structures}

We qualitatively study the proposed reward structure. Since the proposer receives only a scalar validation score without any explicit structural supervision, we investigate whether \textit{it implicitly learn to generate mathematically sound and stable reward formulations}. We classify every reward function proposed in a ALFWorld L2 run into three types based on mathematical properties: \textbf{1) Stable}: linear combinations or normalization mechanisms. For example, linear additions (\textit{e.g.,} $0.5 \cdot g_1 + 0.8 \cdot g_2$) or division operations (\textit{e.g.,} $\frac{g_1}{g_2 + 1}$) that act similarly to sigmoid functions bound the output range. This structure mirrors robust designs in deep learning, preventing numerical explosion while retaining sufficient expressivity to guide the agent. \textbf{2) Unstable}: predominantly unbounded products without normalization. A typical example is a chain of sequential multiplications (\textit{e.g.,} $g_1 \cdot (g_2 + 0.2) \cdot g_3$). This structure creates a severe ``veto'' mechanism: if any single primitive approaches zero, the entire reward vanishes, leading to high variance and unstable gradient updates. \textbf{3) Invalid}: mathematically adversarial forms, such as assigning negative coefficients to positive signals (\textit{e.g.,} $-(g_1 + 0.5 \cdot g_2)$), which penalize desirable behaviors and offer no optimization utility.

\begin{figure}[t]
\centering
\includegraphics[width=\linewidth]{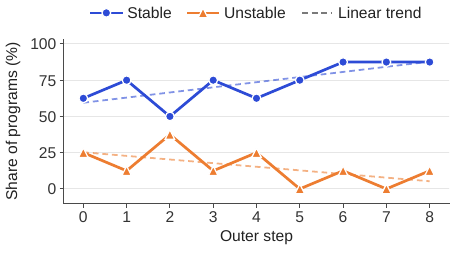}
\caption{Share of stable and unstable structures among eight programs proposed at each outer step of an ALFWorld L2 run (nine steps, dynamics in Appendix~\ref{app:outer_loop_examples}).}
\label{fig:patterns}
\end{figure}

Figure~\ref{fig:patterns} tracks the distribution of these structural types over the outer-loop training on ALFWorld. The optimization trajectory reveals a distinct \textit{gradient-guided refinement}. In early exploration phase, unstable structures appear frequently as the optimizer explores the search space. As training progresses, we observe a sharp decline in their prevalence. Simultaneously, the proportion of stable structures exhibits a strong upward trend, eventually becoming dominant. This dynamic suggests that the outer-loop proposer effectively acts as an evolutionary filter for mathematical robustness.
Without explicit human programming or constraints, \method implicitly discovers that consistent, bounded, and numerically stable rewards are critical prerequisites for effective policy optimization.

\section{Conclusion}
\label{sec:conclusion}


We cast automated reward design for RL as a dual-evolving process, introducing Dual-Evolving Reinforcement Learning (\method). Unlike prior methods with static proposers, \method continuously optimizes a reward proposer via policy gradients derived from the inner policy's validation performance, effectively capturing the empirical meta-gradient of task success. 

Evaluations across embodied, scientific, and mathematical reasoning domains demonstrate that \method synthesizes rewards that train superior policies compared to outcome-based and hand-designed process rewards, particularly on out-of-distribution tasks. Furthermore, compute-matched comparisons confirm these gains stem fundamentally from learning the proposer, while ablation studies highlight the framework's robustness to sub-optimal primitive vocabularies. Ultimately, \method paves the way for scalable reward design with minimal human priors. Future work will explore more computationally efficient inner-loop proxies and fully automated primitive generation.

\section*{Limitations}
One natural limitation of \method is the computational overhead introduced by the outer-loop search compared to a single standard RL run. However, the performance improvements are not merely a result of this extra budget. As shown in our compute-matched evaluations (\S\ref{sec:learned_proposer}) and sample efficiency analysis (Appendix \ref{app:number_of_rollouts}), the learned proposer maintains its advantage even when computational resources are controlled. Additionally, the current framework relies on a human-specified vocabulary of primitives. Fortunately, our sensitivity analysis (\S\ref{sec:sensitivity}) indicates that \method is highly robust to the primitives, meaning it does not depend on perfectly engineered atomic signals to succeed. Future work could further alleviate these constraints by exploring computationally lighter inner-loop proxies and fully automated primitive generation.

\section*{Ethics Statement}

This work automates part of reward design for RL training of language models.
Automatically generated rewards can encode unintended incentives.
The generated programs in this paper are short symbolic expressions over interpretable primitives, which makes them auditable, and we recommend such inspection before deployment.
The outer-loop multiplies training compute and hence energy use relative to a single RL run, and we report the cost so that readers can weigh it.
All benchmarks are public and contain no personal data.

\section*{Use of AI Assistants}

We used AI assistants to help write this paper.
The assistants paraphrased and polished text that we wrote, checked that the text agrees with the tables and figures, and helped with the LaTeX layout.
The authors checked every change and take full responsibility for the content.

\bibliography{example_paper}

@inproceedings{wang2022scienceworld,
  title={{ScienceWorld}: Is your Agent Smarter than a 5th Grader?},
  author={Wang, Ruoyao and Jansen, Peter and C{\^o}t{\'e}, Marc-Alexandre and Ammanabrolu, Prithviraj},
  booktitle={Proceedings of the 2022 Conference on Empirical Methods in Natural Language Processing},
  pages={11279--11298},
  year={2022}
}

@inproceedings{zhuang-etal-2024-efficientrag,
    title = "{E}fficient{RAG}: Efficient Retriever for Multi-Hop Question Answering",
    author = "Zhuang, Ziyuan  and
      Zhang, Zhiyang  and
      Cheng, Sitao  and
      Yang, Fangkai  and
      Liu, Jia  and
      Huang, Shujian  and
      Lin, Qingwei  and
      Rajmohan, Saravan  and
      Zhang, Dongmei  and
      Zhang, Qi",
    editor = "Al-Onaizan, Yaser  and
      Bansal, Mohit  and
      Chen, Yun-Nung",
    booktitle = "Proceedings of the 2024 Conference on Empirical Methods in Natural Language Processing",
    month = nov,
    year = "2024",
    address = "Miami, Florida, USA",
    publisher = "Association for Computational Linguistics",
    url = "https://aclanthology.org/2024.emnlp-main.199/",
    doi = "10.18653/v1/2024.emnlp-main.199",
    pages = "3392--3411",
}

@article{cheng2026xtree,
  title   = {{X-Tree}: Tokenizing Reusable Experience for
             Efficient Agent Generalization},
  author  = {Cheng, Sitao and Yin, Xunjian and Sun, Zhiyuan and
             Li, Yuxuan and Zhou, Ruiwen and Jian, Xiangru and
             Zhong, Victor},
  journal = {arXiv preprint arXiv:2609.32993},
  year    = {2026}
}

@article{hu2025reinforce++,
  title={{REINFORCE++}: An efficient {RLHF} algorithm with robustness to both prompt and reward models},
  author={Hu, Jian and Liu, Jason Klein and Shen, Wei},
  journal={arXiv preprint arXiv:2501.03262},
  year={2025}
}

@article{huang2023markqa,
  title={MarkQA: A large scale KBQA dataset with numerical reasoning},
  author={Huang, Xiang and Cheng, Sitao and Bao, Yuheng and Huang, Shanshan and Qu, Yuzhong},
  journal={arXiv preprint arXiv:2310.15517},
  year={2023}
}

@article{cheng2024understanding,
  title={Understanding the interplay between parametric and contextual knowledge for large language models},
  author={Cheng, Sitao and Pan, Liangming and Yin, Xunjian and Wang, Xinyi and Wang, William Yang},
  journal={arXiv preprint arXiv:2410.08414},
  year={2024}
}

@article{cheng2025atomic,
  title={From Atomic to Composite: Reinforcement Learning Enables Generalization in Complementary Reasoning},
  author={Cheng, Sitao and Yin, Xunjian and Zhou, Ruiwen and Li, Yuxuan and Wang, Xinyi and Pan, Liangming and Wang, William Yang and Zhong, Victor},
  journal={arXiv preprint arXiv:2512.01970},
  year={2025}
}

@article{zhang2025rlvmr,
  title={{RLVMR}: Reinforcement learning with verifiable meta-reasoning rewards for robust long-horizon agents},
  author={Zhang, Zijing and Chen, Ziyang and Li, Mingxiao and Tu, Zhaopeng and Li, Xiaolong},
  journal={arXiv preprint arXiv:2507.22844},
  year={2025}
}

@article{cobbe2021gsm8k,
  title={Training Verifiers to Solve Math Word Problems},
  author={Cobbe, Karl and Kosaraju, Vineet and Bavarian, Mohammad and Chen, Mark and Jun, Heewoo and Kaiser, Lukasz and Plappert, Matthias and Tworek, Jerry and Hilton, Jacob and Nakano, Reiichiro and Hesse, Christopher and Schulman, John},
  journal={arXiv preprint arXiv:2110.14168},
  year={2021}
}

@inproceedings{andrychowicz2016learninglearngradientdescent,
  title={Learning to Learn by Gradient Descent by Gradient Descent},
  author={Marcin Andrychowicz and Misha Denil and Sergio Gomez and Matthew W. Hoffman and David Pfau and Tom Schaul and Brendan Shillingford and Nando de Freitas},
  booktitle={Advances in Neural Information Processing Systems},
  volume={29},
  year={2016}
}

@article{schulman2017proximal,
  title={Proximal policy optimization algorithms},
  author={Schulman, John and Wolski, Filip and Dhariwal, Prafulla and Radford, Alec and Klimov, Oleg},
  journal={arXiv preprint arXiv:1707.06347},
  year={2017}
}

@article{vilalta2002meta-survey,
  title={A Perspective View and Survey of Meta-Learning},
  author={Vilalta, Ricardo and Drissi, Youssef},
  journal={Artificial Intelligence Review},
  volume={18},
  number={2},
  pages={77--95},
  year={2002},
  publisher={Springer}
}

@inproceedings{qin2023toolllm,
  title={{ToolLLM}: Facilitating Large Language Models to Master 16000+ Real-world {APIs}},
  author={Qin, Yujia and Liang, Shihao and Ye, Yining and Zhu, Kunlun and Yan, Lan and Lu, Yaxi and Lin, Yankai and Cong, Xin and Tang, Xiangru and Qian, Bill and others},
  booktitle={The Twelfth International Conference on Learning Representations},
  year={2024}
}

@article{jaderberg2017population,
  title={Population based training of neural networks},
  author={Jaderberg, Max and Dalibard, Valentin and Osindero, Simon and Czarnecki, Wojciech M and Donahue, Jeff and Razavi, Ali and Vinyals, Oriol and Green, Tim and Dunning, Iain and Simonyan, Karen and others},
  journal={arXiv preprint arXiv:1711.09846},
  year={2017}
}

@inproceedings{yu2025dapo,
  title={{DAPO}: An Open-Source {LLM} Reinforcement Learning System at Scale},
  author={Yu, Qiying and Zhang, Zheng and Zhu, Ruofei and Yuan, Yufeng and Zuo, Xiaochen and Yue, Yu and Dai, Weinan and Fan, Tiantian and Liu, Gaohong and Liu, Lingjun and others},
  booktitle={Advances in Neural Information Processing Systems},
  volume={38},
  year={2025}
}

@inproceedings{knox2009interactively,
  title={Interactively shaping agents via human reinforcement: The TAMER framework},
  author={Knox, W Bradley and Stone, Peter},
  booktitle={Proceedings of the fifth international conference on Knowledge capture},
  pages={9--16},
  year={2009}
}

@article{gao2025survey,
  title={A survey of self-evolving agents: On path to artificial super intelligence},
  author={Gao, Huan-ang and Geng, Jiayi and Hua, Wenyue and Hu, Mengkang and Juan, Xinzhe and Liu, Hongzhang and Liu, Shilong and Qiu, Jiahao and Qi, Xuan and Wu, Yiran and others},
  journal={arXiv preprint arXiv:2507.21046},
  year={2025}
}

@article{zhang2025darwin,
  title={{Darwin G{\"o}del Machine}: Open-Ended Evolution of Self-Improving Agents},
  author={Zhang, Jenny and Hu, Shengran and Lu, Cong and Lange, Robert and Clune, Jeff},
  journal={arXiv preprint arXiv:2505.22954},
  year={2025}
}

@misc{sutton2019bitter,
  title={The Bitter Lesson},
  author={Sutton, Richard},
  howpublished={Incomplete Ideas (blog)},
  year={2019}
}

@inproceedings{shridhar2020alfworld,
  title={{ALFWorld}: Aligning Text and Embodied Environments for Interactive Learning},
  author={Shridhar, Mohit and Yuan, Xingdi and C{\^o}t{\'e}, Marc-Alexandre and Bisk, Yonatan and Trischler, Adam and Hausknecht, Matthew},
  booktitle={International Conference on Learning Representations},
  year={2021}
}

@article{sheng2024verl,
  title   = {HybridFlow: A Flexible and Efficient {RLHF} Framework},
  author  = {Guangming Sheng and Chi Zhang and Zilingfeng Ye and Xibin Wu and Wang Zhang and Ru Zhang and Yanghua Peng and Haibin Lin and Chuan Wu},
  year    = {2024},
  journal = {arXiv preprint arXiv: 2409.19256}
}

@inproceedings{hendrycks2021measuring,
  title={Measuring Mathematical Problem Solving with the {MATH} Dataset},
  author={Hendrycks, Dan and Burns, Collin and Kadavath, Saurav and Arora, Akul and Basart, Steven and Tang, Eric and Song, Dawn and Steinhardt, Jacob},
  booktitle={Advances in Neural Information Processing Systems: Datasets and Benchmarks Track},
  year={2021}
}

@article{shao2024deepseekmath,
  title={Deepseekmath: Pushing the limits of mathematical reasoning in open language models},
  author={Shao, Zhihong and Wang, Peiyi and Zhu, Qihao and Xu, Runxin and Song, Junxiao and Bi, Xiao and Zhang, Haowei and Zhang, Mingchuan and Li, YK and Wu, Yang and others},
  journal={arXiv preprint arXiv:2402.03300},
  year={2024}
}

@article{deepseekai2025deepseekr1incentivizingreasoningcapability,
  title={{DeepSeek-R1} incentivizes reasoning in {LLMs} through reinforcement learning},
  author={Guo, Daya and Yang, Dejian and Zhang, Haowei and others},
  journal={Nature},
  volume={645},
  number={8081},
  pages={633--638},
  year={2025},
  doi={10.1038/s41586-025-09422-z}
}

@inproceedings{agarwal2019learning,
  title={Learning to Generalize from Sparse and Underspecified Rewards},
  author={Agarwal, Rishabh and Liang, Chen and Schuurmans, Dale and Norouzi, Mohammad},
  booktitle={Proceedings of the 36th International Conference on Machine Learning},
  pages={130--140},
  year={2019},
  publisher={PMLR}
}

@article{amodei2016concrete,
  title={Concrete problems in {AI} safety},
  author={Amodei, Dario and Olah, Chris and Steinhardt, Jacob and Christiano, Paul and Schulman, John and Man{\'e}, Dan},
  journal={arXiv preprint arXiv:1606.06565},
  year={2016}
}

@inproceedings{feng2025group,
  title={Group-in-Group Policy Optimization for {LLM} Agent Training},
  author={Feng, Lang and Xue, Zhenghai and Liu, Tingcong and An, Bo},
  booktitle={Advances in Neural Information Processing Systems},
  year={2025}
}

@article{shao2025dr,
  title={{DR Tulu}: Reinforcement Learning with Evolving Rubrics for Deep Research},
  author={Shao, Rulin and Asai, Akari and Shen, Shannon Zejiang and Ivison, Hamish and Kishore, Varsha and Zhuo, Jingming and Zhao, Xinran and Park, Molly and Finlayson, Samuel G and Sontag, David and others},
  journal={arXiv preprint arXiv:2511.19399},
  year={2025}
}

@article{such2017deep,
  title={Deep neuroevolution: Genetic algorithms are a competitive alternative for training deep neural networks for reinforcement learning},
  author={Such, Felipe Petroski and Madhavan, Vashisht and Conti, Edoardo and Lehman, Joel and Stanley, Kenneth O and Clune, Jeff},
  journal={arXiv preprint arXiv:1712.06567},
  year={2017}
}

@article{skalse2022defining,
  title={Defining and characterizing reward gaming},
  author={Skalse, Joar and Howe, Nikolaus and Krasheninnikov, Dmitrii and Krueger, David},
  journal={Advances in Neural Information Processing Systems},
  volume={35},
  pages={9460--9471},
  year={2022}
}

@article{oh2020discovering,
  title={Discovering reinforcement learning algorithms},
  author={Oh, Junhyuk and Hessel, Matteo and Czarnecki, Wojciech M and Xu, Zhongwen and van Hasselt, Hado P and Singh, Satinder and Silver, David},
  journal={Advances in Neural Information Processing Systems},
  volume={33},
  pages={1060--1070},
  year={2020}
}

@inproceedings{cheng-etal-2024-call,
    title = "Call Me When Necessary: {LLM}s can Efficiently and Faithfully Reason over Structured Environments",
    author = "Cheng, Sitao  and
      Zhuang, Ziyuan  and
      Xu, Yong  and
      Yang, Fangkai  and
      Zhang, Chaoyun  and
      Qin, Xiaoting  and
      Huang, Xiang  and
      Chen, Ling  and
      Lin, Qingwei  and
      Zhang, Dongmei  and
      Rajmohan, Saravan  and
      Zhang, Qi",
    editor = "Ku, Lun-Wei  and
      Martins, Andre  and
      Srikumar, Vivek",
    booktitle = "Findings of the Association for Computational Linguistics: ACL 2024",
    month = aug,
    year = "2024",
    address = "Bangkok, Thailand",
    publisher = "Association for Computational Linguistics",
    url = "https://aclanthology.org/2024.findings-acl.254/",
    doi = "10.18653/v1/2024.findings-acl.254",
    pages = "4275--4295",
}

@inproceedings{huang-etal-2024-queryagent,
    title = "{Q}uery{A}gent: A Reliable and Efficient Reasoning Framework with Environmental Feedback based Self-Correction",
    author = "Huang, Xiang  and
      Cheng, Sitao  and
      Huang, Shanshan  and
      Shen, Jiayu  and
      Xu, Yong  and
      Zhang, Chaoyun  and
      Qu, Yuzhong",
    editor = "Ku, Lun-Wei  and
      Martins, Andre  and
      Srikumar, Vivek",
    booktitle = "Proceedings of the 62nd Annual Meeting of the Association for Computational Linguistics (Volume 1: Long Papers)",
    month = aug,
    year = "2024",
    address = "Bangkok, Thailand",
    publisher = "Association for Computational Linguistics",
    url = "https://aclanthology.org/2024.acl-long.274/",
    doi = "10.18653/v1/2024.acl-long.274",
    pages = "5014--5035",
}

@inproceedings{an-etal-2025-thread,
    title = "Thread: A Logic-Based Data Organization Paradigm for How-To Question Answering with Retrieval Augmented Generation",
    author = "An, Kaikai  and
      Yang, Fangkai  and
      Li, Liqun  and
      Lu, Junting  and
      Cheng, Sitao  and
      Si, Shuzheng  and
      Wang, Lu  and
      Zhao, Pu  and
      Cao, Lele  and
      Lin, Qingwei  and
      Rajmohan, Saravan  and
      Zhang, Dongmei  and
      Chang, Baobao",
    editor = "Christodoulopoulos, Christos  and
      Chakraborty, Tanmoy  and
      Rose, Carolyn  and
      Peng, Violet",
    booktitle = "Proceedings of the 2025 Conference on Empirical Methods in Natural Language Processing",
    month = nov,
    year = "2025",
    address = "Suzhou, China",
    publisher = "Association for Computational Linguistics",
    url = "https://aclanthology.org/2025.emnlp-main.923/",
    doi = "10.18653/v1/2025.emnlp-main.923",
    pages = "18300--18319",
    ISBN = "979-8-89176-332-6",
}

@article{ouyang2022training,
  title={Training language models to follow instructions with human feedback},
  author={Ouyang, Long and Wu, Jeffrey and Jiang, Xu and Almeida, Diogo and Wainwright, Carroll and Mishkin, Pamela and Zhang, Chong and Agarwal, Sandhini and Slama, Katarina and Ray, Alex and others},
  journal={Advances in neural information processing systems},
  volume={35},
  pages={27730--27744},
  year={2022}
}

@article{xu2018meta,
  title={Meta-gradient reinforcement learning},
  author={Xu, Zhongwen and van Hasselt, Hado P and Silver, David},
  journal={Advances in neural information processing systems},
  volume={31},
  year={2018}
}

@article{novikov2025alphaevolve,
  title={AlphaEvolve: A coding agent for scientific and algorithmic discovery},
  author={Novikov, Alexander and V{\~u}, Ng{\^a}n and Eisenberger, Marvin and Dupont, Emilien and Huang, Po-Sen and Wagner, Adam Zsolt and Shirobokov, Sergey and Kozlovskii, Borislav and Ruiz, Francisco JR and Mehrabian, Abbas and others},
  journal={arXiv preprint arXiv:2506.13131},
  year={2025}
}

@misc{yan2025reformreducinghuman,
      title={{Re:Form} -- Reducing Human Priors in Scalable Formal Software Verification with {RL} in {LLMs}: A Preliminary Study on {Dafny}}, 
      author={Chuanhao Yan and Fengdi Che and Xuhan Huang and Xu Xu and Xin Li and Yizhi Li and Xingwei Qu and Jingzhe Shi and Chenghua Lin and Yaodong Yang and Binhang Yuan and Hang Zhao and Yu Qiao and Bowen Zhou and Jie Fu},
      year={2025},
      eprint={2507.16331},
      archivePrefix={arXiv},
      primaryClass={cs.CL},
      url={https://arxiv.org/abs/2507.16331}, 
}

@inproceedings{Bello2017Neural,
  author       = {Irwan Bello and
                  Barret Zoph and
                  Vijay Vasudevan and
                  Quoc V. Le},
  editor       = {Doina Precup and
                  Yee Whye Teh},
  title        = {Neural Optimizer Search with Reinforcement Learning},
  booktitle    = {Proceedings of the 34th International Conference on Machine Learning,
                  {ICML} 2017, Sydney, NSW, Australia, 6-11 August 2017},
  series       = {Proceedings of Machine Learning Research},
  volume       = {70},
  pages        = {459--468},
  publisher    = {{PMLR}},
  year         = {2017},
  url          = {http://proceedings.mlr.press/v70/bello17a.html},
  bibsource    = {dblp computer science bibliography, https://dblp.org}
}

@inproceedings{Ma2024eureka,
  author       = {Yecheng Jason Ma and
                  William Liang and
                  Guanzhi Wang and
                  De{-}An Huang and
                  Osbert Bastani and
                  Dinesh Jayaraman and
                  Yuke Zhu and
                  Linxi Fan and
                  Anima Anandkumar},
  title        = {Eureka: Human-Level Reward Design via Coding Large Language Models},
  booktitle    = {The Twelfth International Conference on Learning Representations,
                  {ICLR} 2024, Vienna, Austria, May 7-11, 2024},
  publisher    = {OpenReview.net},
  year         = {2024},
  url          = {https://openreview.net/forum?id=IEduRUO55F},
  bibsource    = {dblp computer science bibliography, https://dblp.org}
}

@article{Romera24mathematical,
  author       = {Bernardino Romera{-}Paredes and
                  Mohammadamin Barekatain and
                  Alexander Novikov and
                  Matej Balog and
                  M. Pawan Kumar and
                  Emilien Dupont and
                  Francisco J. R. Ruiz and
                  Jordan S. Ellenberg and
                  Pengming Wang and
                  Omar Fawzi and
                  Pushmeet Kohli and
                  Alhussein Fawzi},
  title        = {Mathematical discoveries from program search with large language models},
  journal      = {Nature},
  volume       = {625},
  number       = {7995},
  pages        = {468--475},
  year         = {2024},
  url          = {https://doi.org/10.1038/s41586-023-06924-6},
  doi          = {10.1038/S41586-023-06924-6},
  bibsource    = {dblp computer science bibliography, https://dblp.org}
}

@misc{tang2025calmstormunlockingnative,
      title={{CALM} Before the {STORM}: Unlocking Native Reasoning for Optimization Modeling},
      author={Zhengyang Tang and Zihan Ye and Chenyu Huang and Xuhan Huang and Chengpeng Li and Sihang Li and Guanhua Chen and Ming Yan and Zizhuo Wang and Hongyuan Zha and Dayiheng Liu and Benyou Wang},
      year={2025},
      eprint={2510.04204},
      archivePrefix={arXiv},
      primaryClass={cs.CL},
      url={https://arxiv.org/abs/2510.04204},
}

@article{niekum2010genetic,
  title={Genetic programming for reward function search},
  author={Niekum, Scott and Barto, Andrew G. and Spector, Lee},
  journal={IEEE Transactions on Autonomous Mental Development},
  volume={2},
  number={2},
  pages={83--90},
  year={2010}
}

@article{faust2019evolving,
  title={Evolving rewards to automate reinforcement learning},
  author={Faust, Aleksandra and Francis, Anthony and Mehta, Dar},
  journal={arXiv preprint arXiv:1905.07628},
  year={2019}
}

@inproceedings{khadka2018evolution,
  title={Evolution-guided policy gradient in reinforcement learning},
  author={Khadka, Shauharda and Tumer, Kagan},
  booktitle={Advances in Neural Information Processing Systems},
  volume={31},
  year={2018}
}

@inproceedings{zoph2017neural,
  title={Neural architecture search with reinforcement learning},
  author={Zoph, Barret and Le, Quoc V.},
  booktitle={International Conference on Learning Representations},
  year={2017}
}

@article{williams1992simple,
  title={Simple statistical gradient-following algorithms for connectionist reinforcement learning},
  author={Williams, Ronald J.},
  journal={Machine Learning},
  volume={8},
  number={3--4},
  pages={229--256},
  year={1992}
}

@inproceedings{singh2009where,
  title={Where do rewards come from?},
  author={Singh, Satinder and Lewis, Richard L. and Barto, Andrew G.},
  booktitle={Proceedings of the Annual Conference of the Cognitive Science Society},
  pages={2601--2606},
  year={2009}
}

@inproceedings{zheng2018learning,
  title={On learning intrinsic rewards for policy gradient methods},
  author={Zheng, Zeyu and Oh, Junhyuk and Singh, Satinder},
  booktitle={Advances in Neural Information Processing Systems},
  volume={31},
  year={2018}
}

@inproceedings{lee2023rlaif,
  title={{RLAIF} vs. {RLHF}: Scaling Reinforcement Learning from Human Feedback with {AI} Feedback},
  author={Lee, Harrison and Phatale, Samrat and Mansoor, Hassan and Mesnard, Thomas and Ferret, Johan and Lu, Kellie and Bishop, Colton and Hall, Ethan and Carbune, Victor and Rastogi, Abhinav and Prakash, Sushant},
  booktitle={Proceedings of the 41st International Conference on Machine Learning},
  year={2024}
}

@article{surina2025algorithm,
  title={Algorithm Discovery With {LLMs}: Evolutionary Search Meets Reinforcement Learning},
  author={Surina, Anja and Mansouri, Amin and Quaedvlieg, Lars and others},
  journal={arXiv preprint arXiv:2504.05108},
  year={2025}
}

@article{zha2025rltango,
  title={{RL Tango}: Reinforcing Generator and Verifier Together for Language Reasoning},
  author={Zha, Kaiwen and Gao, Zhengqi and Shen, Maohao and others},
  journal={arXiv preprint arXiv:2505.15034},
  year={2025}
}

@article{gunjal2025rubrics,
  title={Rubrics as Rewards: Reinforcement Learning Beyond Verifiable Domains},
  author={Gunjal, Anisha and Wang, Anthony and Lau, Elaine and others},
  journal={arXiv preprint arXiv:2507.17746},
  year={2025}
}

@article{cui2025process,
  title={Process Reinforcement through Implicit Rewards},
  author={Cui, Ganqu and Yuan, Lifan and Wang, Zefan and others},
  journal={arXiv preprint arXiv:2502.01456},
  year={2025}
}

@inproceedings{xie2024text2reward,
  title={{Text2Reward}: Reward Shaping with Language Models for Reinforcement Learning},
  author={Xie, Tianbao and Zhao, Siheng and Wu, Chen Henry and others},
  booktitle={International Conference on Learning Representations},
  year={2024}
}

@inproceedings{song2024trial,
  title={Trial and Error: Exploration-Based Trajectory Optimization for {LLM} Agents},
  author={Song, Yifan and Yin, Da and Yue, Xiang and Huang, Jie and Li, Sujian and Lin, Bill Yuchen},
  booktitle={Proceedings of the 62nd Annual Meeting of the Association for Computational Linguistics (Volume 1: Long Papers)},
  year={2024}
}

@inproceedings{xiong2024watch,
  title={Watch Every Step! {LLM} Agent Learning via Iterative Step-Level Process Refinement},
  author={Xiong, Weimin and Song, Yifan and Zhao, Xiutian and others},
  booktitle={Proceedings of the 2024 Conference on Empirical Methods in Natural Language Processing},
  year={2024}
}

@inproceedings{lightman2024lets,
  title={Let's Verify Step by Step},
  author={Lightman, Hunter and Kosaraju, Vineet and Burda, Yuri and Edwards, Harrison and Baker, Bowen and Lee, Teddy and Leike, Jan and Schulman, John and Sutskever, Ilya and Cobbe, Karl},
  booktitle={International Conference on Learning Representations},
  year={2024}
}

@inproceedings{dang2026temperature,
  title={Temperature as a Meta-Policy: Adaptive Temperature in {LLM} Reinforcement Learning},
  author={Dang, Haoran and Lan, Cuiling and Wan, Hai and Zhao, Xibin and Lu, Yan},
  booktitle={The Fourteenth International Conference on Learning Representations},
  year={2026}
}

\appendix

\section{Full Procedure of \method}
\label{app:algorithm}

Algorithm~\ref{alg:derl} lists the full procedure of \method.
Lines 1--9 are the dual-evolving training.
Lines 10--11 are the final training run.

\begin{algorithm}[ht]
\caption{\method}
\label{alg:derl}
\small
\begin{algorithmic}[1]
\REQUIRE instruction $\ins$, proposer $\pi_\psi$, base policy $\theta_0$, splits $\din$ and $\dev$, rollout size $n$, outer steps $E_{\mathrm{out}}$
\FOR{outer step $t = 1,\dots,E_{\mathrm{out}}$}
    \STATE sample $\phi_1,\dots,\phi_n \sim \pi_\psi(\cdot \mid \ins)$ under constrained decoding (Appendix~\ref{app:exp_protocol})
    \FOR{$i = 1,\dots,n$ in parallel}
        \STATE $\theta_i \leftarrow$ GRPO on $\din$ with reward $R_{\phi_i}$ (Eqs.~\ref{eq:inner:adv}--\ref{eq:grpo}), from $\theta_0$ (\methodbase) or from the best checkpoint of step $t{-}1$ (\methodfull)
        \STATE $v_i \leftarrow \Perf(\theta_i;\dev)$ (Eq.~\ref{eq:perf}), or $v_i \leftarrow 0$ if $\phi_i$ raises an execution error
    \ENDFOR
    \STATE $\hat{A}_i \leftarrow (v_i - \bar{v}) / \sigma_v$ for $i=1,\dots,n$
    \STATE update $\psi$ with the GRPO objective, $\hat{A}_i$ being the advantage of every token of $\phi_i$ (estimates Eq.~\ref{eq:outer:gradient})
\ENDFOR
\STATE $\phi^\star \leftarrow$ the found reward function
\STATE train a fresh policy on the full training set under the baseline budget, with $R_{\phi^\star}$ (\methodbase) or with the same set of found rewards (\methodfull), and evaluate it once on the test set (\S\ref{sec:protocol})
\end{algorithmic}
\end{algorithm}

\section{Primitive Design and Examples}
\label{app:primitive_design}

\subsection{Properties of the Reward Structure}

\paragraph{Coverage.}
The expression space of Equation~\eqref{reward_fun} contains the binary outcome reward $\phi = g_1$ and the uniform average $\phi = (g_1+\dots+g_k)/k$, so the search space covers the outcome baselines.

\paragraph{Structured actions.}
The proposer decides which signals to combine and how, while the primitives parse the policy outputs.

\paragraph{Extensibility.}
Primitive definitions are decoupled from their use.
A new task needs only new primitive code. And the outer-loop is robust to  primitives not helpful (\S\ref{sec:sensitivity}).

\subsection{Primitives Used in This Paper}
\label{app:primitive_examples}

\paragraph{Agents.}
Outcome rewards arrive at the end of long trajectories \cite{zhang2025rlvmr,cheng2025atomic}.
To give the outer-loop building blocks for dense credit assignment, we define primitives over temporal phases of the trajectory.
Following RLVMR~\citep{zhang2025rlvmr}, the policy tags each action with a meta-tag (exploration, planning, or formatting), and a rule-based string match between the action and its tag yields a step score.
We divide the trajectory evenly into three phases and define four primitives:
\begin{itemize}[leftmargin=*,itemsep=1pt,topsep=0pt]
    \item $g_1 = 1$ if the task is completed, else $0$.
    \item $g_2$ = mean step score over the first third.
    \item $g_3$ = mean step score over the middle third.
    \item $g_4$ = mean step score over the last third.
\end{itemize}
Example program found by \method:
\[
g_1 + (g_2 \cdot (g_3 / 2)) - (g_4 / 3).
\]

\paragraph{Mathematics.}
Let $o$ be the raw output, $\hat{a}$ the extracted answer, and $a^*$ the ground truth.
\begin{itemize}[leftmargin=*,itemsep=1pt,topsep=0pt]
    \item $g_1 = 1$ if $\hat{a} = a^*$, else $0$.
    \item $g_2 = 1$ if $\hat{a}$ is enclosed in \verb|\boxed{}|, else $0$.
    \item $g_3 = 1$ if step markers such as ``Step 1'' appear in $o$, else $0$.
    \item $g_4 = 1$ if $a^\ast$ appears in $o$ (soft match), else $0$.
\end{itemize}
Example program found by \method:
\[
g_1 + 0.5 \cdot g_2 + 0.1 \cdot (g_3 + g_4).
\]

\section{Experimental Protocol and Settings}
\label{app:exp_protocol}

\paragraph{Data split.}
Let $\dtrain$ be the original training set.
During reward search we split it at random 8:2 into $\din$ and $\dev$.
Each configuration $\phi_i$ trains an inner policy $\theta_i$ on $\din$.
We use $\dev$ only to compute $v_i=\Perf(\theta_i;\dev)$, which updates the proposer and selects the final configuration.
Reward search never uses the test set $\dtest$.

\paragraph{Final training and evaluation.}
After reward search, the search-stage policies are discarded. For \methodbase, a fresh policy is trained on the full $\dtrain$ with the selected reward. For \methodfull, a fresh policy is trained on the full $\dtrain$ using the reward program associated with the selected policy at each search round, in the original round order and with the same per-round training budgets. Training continues from the preceding round's checkpoint when the reward changes.
We evaluate this policy once on $\dtest$.
It receives the same data and no more training epochs than the baselines.
The extra cost of \method is the reward search (Appendix~\ref{app:cost_ana}).

\paragraph{Outer-loop settings.}
\texttt{Qwen2.5-0.5B-Instruct} is the proposer, trained with GRPO under the default VeRL settings~\citep{sheng2024verl}.
During RL rollouts a logit mask restricts decoding to the primitive names $g_1,\dots,g_k$, the operators $+$, $-$, $*$, $/$, and $**$ (exponentiation), digits and the decimal point, and parentheses.
The mask does not enforce balanced parentheses.
A generated configuration that raises an execution error receives $v_i = 0$.
All experiments by default use the rollout size $n=8$.

\paragraph{Inner-loop settings.}
For ALFWorld and ScienceWorld the base policy is \texttt{Qwen2.5-1.5B-Instruct}, shared with the baselines.
We count budgets in the epoch unit of RLVMR.
Inner runs during the search last 40 epochs on ALFWorld and 80 on ScienceWorld. The final policy trains from the base policy for 100 epochs.
\methodfull runs 10 rounds of 10 epochs on ALFWorld and 3 rounds of 33 epochs on ScienceWorld, and its reported policy is trained on the full training set with the same set of found rewards.
Evaluation follows the setting of RLVMR~\citep{zhang2025rlvmr}.
For mathematics the base policy is \texttt{Qwen2.5-3B}.
Inner runs during the search last 10 epochs with a 3.5-hour cap, and the final policy is trained for 15 epochs.
\methodfull trains 2 inner-training epochs per round, and its reported performance is measured after the seventh round of final training on the full training set.

\paragraph{Generalization levels.}
\label{app:generalization_settings}
We test models capabilities with varied difficulties \cite{huang2023markqa,cheng2024understanding}.
ALFWorld has six task categories (\textit{e.g.,} Pick \& Place, Examine in Light).
We train either on all six (in-distribution) or on four (out-of-distribution).
The test set has two splits:
\begin{itemize}[leftmargin=*,itemsep=1pt,topsep=1pt]
    \item \textbf{Seen}: the same object and receptacle concepts as in training, in new positions or quantities (a pencil seen in training appears in a new location).
    \item \textbf{Unseen}: object variants and instructions never seen in training (``put a watch in the safe'' when neither appeared in training).
\end{itemize}
The three levels are \textbf{L0} (in-\hspace{0pt}distribution, seen), \textbf{L1} (in-\hspace{0pt}distribution, unseen), and \textbf{L2} (out-of-\hspace{0pt}distribution, unseen). \citet{zhang2025rlvmr} define the ScienceWorld levels in the same way.

\section{Training Dynamics of \methodfull}
\label{app:training_dynamics_of_pop}

\begin{figure}[t]
    \centering
    \includegraphics[width=\linewidth]{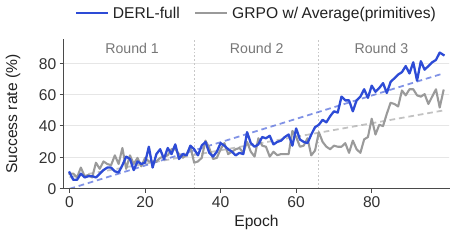}
    \vspace{-22pt}
    \caption{Success rate over 100 training epochs on ScienceWorld L0 for \methodfull (Three rounds of 33 epochs. Each round stars from the best inner-policy of the previous one) and GRPO w/ Average(primitives).}
    \vspace{-10pt}
    \label{fig:DERL-pop}
\end{figure}

Figure~\ref{fig:DERL-pop} compares \methodfull with GRPO w/ Average(primitives) on ScienceWorld L0 under the same budget.
In the first round, the two curves overlap, since both policies start from the base model and the outer-loop has not yet updated.
From the second round on, \methodfull trains under a reward optimized for the current inner policy and pulls away, and in the third round the gap widens further.
In \methodfull, the reward  is optimized for the current round of training, and each round starts from the best checkpoint of the previous round. The reported \methodfull policy (Table~\ref{tab:results_on_agent}) involves no such selection, because its final training replays the selected rewards in one run (Appendix~\ref{app:exp_protocol}).

\section{Examples Programs from the Proposer}
\label{app:outer_loop_examples}

\begin{figure}[t]
\centering
\includegraphics[width=\linewidth]{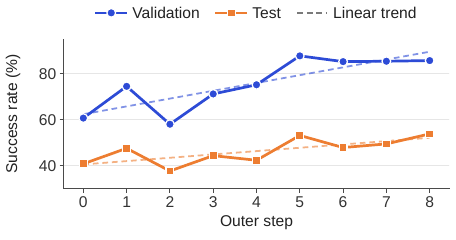}
\vspace{-20pt}
\caption{Outer-loop dynamics of the ALFWorld L2 run analyzed in Figure~\ref{fig:patterns}, which converges in nine steps. Axes and curves are as in Figure~\ref{fig:analysis}.}
\vspace{-10pt}
\label{fig:ppo2_dynamics}
\end{figure}

\begin{table*}[t]
    \centering
    \footnotesize
    \setlength{\tabcolsep}{5pt}
    \begin{tabular}{cl@{\hspace{10pt}}c}
        \toprule
        \textbf{Outer step} & \textbf{Proposed program (verbatim)} & \textbf{Val score} \\
        \midrule
        \multirow{8}{*}{0} & \texttt{g1 * (g2 - 1) / 2 + (g3 + 1) * (g4 - 1) * 2 / 3} & 0 \\
         & \texttt{g1 + 0.5 * (g2 + 0.5 * (g3 + 0.5 * (g4)))) - 0.5 * (g2 + 0.5 * (g3 + ...} & 0 \\
         & \texttt{g1 + 0.01 * (g2 - 0.001) + 0.0001 * (g3 - 0.0001) + 0.000005 * (g4 - ...} & 0.850 \\
         & \texttt{g1 * 0.5 + 0.2 * (g2 + 0.1) - 0.3 / 2 + 0.4 * (g3 * 0.1) + 0.1 * (g4 * ...} & 0.879 \\
         & \texttt{g1 * (g2 + (g3 - 1.0) * (g4 - 0.0))} & 0.023 \\
         & \texttt{g1 + 0.5 * (g2 + 0.5 * (g3 + 0.5 * (g4 + 0.5)))} & 0.885 \\
         & \texttt{g1 + 0.5 * (g2 + 0.2 * (g3 + 0.1 * (g4 + 0.05))))} & 0 \\
         & \texttt{g1 * (g2 + 2 * (g3 / 3)) + (g4 - 4 * (g2 - 1)) - 2.0} & 0.895 \\
        \midrule
        \multirow{8}{*}{1} & \texttt{(0.5 * (g1 - 0.1)) + (0.5 * (g2 - 0.1)) + (0.5 * (g3 - 0.1)) + (0.5 * ...} & 0.779 \\
         & \texttt{g1 * (g2 + (g3 * (g4 - (g2 + (g3 * (g4 - (g3 * (g4 - (g3 * (g4 - (g3 * ...} & 0 \\
         & \texttt{- (g1 + 0.5 * (g2 + 0.3 * (g3 - 0.2 * (g4 + 0.1 * 1))) + 0.1 * 1) / 1.2} & 0.002 \\
         & \texttt{g1 * (g2 - 1) * (1 - 0.5) + 0.5 ** 2 * (g3 - 1) * (1 - 0.25) + 0.25 ** ...} & 0 \\
         & \texttt{g1 + (g2 * (g3 / 2)) - (g4 / 3)} & 0.859 \\
         & \texttt{g1 * (g2 + 0.5) + 0.2 * (g3 * 0.3 + 0.1) - 0.1 * (g4 * 0.2 + 0.5)} & 0.848 \\
         & \texttt{g1 + 2 * (g2 + 3 * (g3 - 2)) * 0.1 + 4 * 0.3 - 5 * (g4 + 1)} & 0.898 \\
         & \texttt{g1 * (g2 - 1) / 2 + (g3 - 1) / 3 + (g4 - 1) / 4 + 3.0} & 0.010 \\
        \midrule
        \multirow{8}{*}{2} & \texttt{g1 + 0.5 * (g2 + 0.5 * (g3 + 0.5 * (g4))))} & 0 \\
         & \texttt{g1 + 0.5 * (g2 - 0.5) + 0.3 * (g3 - 0.05) + 0.2 * (g4 + 0.05) - 0.2 * ...} & 0 \\
         & \texttt{g1 + (g2 * (g3 / 2)) - (g4 / 4)} & 0.844 \\
         & \texttt{g1 + 0.5 * (g2 + 0.5 * (g3 + 0.5 * (g4 - 1))))} & 0 \\
         & \texttt{g1 + 0.01 + 0.0001 * (g2) + 0.000001 * (g3) + 0.00000001 * (g4)} & 0.865 \\
         & \texttt{g1 * (g2 + (g3 * (g4 / 2))) + (g1 - (g2 + (g3 * (g4 / 2)))) * 0.5} & 0.887 \\
         & \texttt{g1 * 0.99 + 0.01 * (g2 + 0.99) + 0.005 * (g3 + 0.99) + 0.0005 * (g4 + ...} & 0.848 \\
         & \texttt{g1 + 0.5 * (g2 - 0.5) + 0.2 * (g3 - 0.5) + 0.1 * (g4 - 0.5)} & 0.891 \\
        \midrule
        \multirow{8}{*}{3} & \texttt{g1 + (g2 / 2.0) - (g3 * 0.1) + (g4 * 0.05)} & 0.844 \\
         & \texttt{g1 + 0.5 * (g2 + 0.5 * (g3 + 0.5 * (g4 + 0.5)))} & 0.875 \\
         & \texttt{g1 + 0.05 * (g2 + 0.05 * (g3 + 0.05 * (g4 + 0.05)))} & 0.824 \\
         & \texttt{g1 + 0.5 * (g2 / 2.0) + 0.1 * (g3 * 2.0) + 0.25 * (g4 / 4.0)} & 0.863 \\
         & \texttt{g1 + 0.5 * (g2 + 0.4 * (g3 + 0.2 * (g4 + 0.1)))} & 0.850 \\
         & \texttt{g1 + 0.5 * (g2 + 0.5 * (g3 + 0.5 * (g4 + 0.5)))} & 0.893 \\
         & \texttt{g1 + 0.5 * (g2 + 0.5 * (g3 + 0.5 * (g4 + 0.5)))} & 0.879 \\
         & \texttt{g1 + (g2 * (g3 / 2)) - (g4 / 3)} & 0.898 \\
        \bottomrule
    \end{tabular}
    \caption{Programs proposed in the first four outer steps of one ALFWorld L2 run, with the validation performance of the inner policy trained with each. Rows with \texttt{...} are cut for display. The text explains the two sources of a 0 score.}
    \vspace{-8pt}
    \label{tab:training_dynamic_example}
\end{table*}

Figure~\ref{fig:ppo2_dynamics} shows the outer-loop dynamics of the run behind Figure~\ref{fig:patterns}.
Table~\ref{tab:training_dynamic_example} lists the examples of programs proposed at each of the first four outer steps of one ALFWorld L2 run (a different run from Figures~\ref{fig:analysis} and \ref{fig:patterns}), with the validation score of the inner policy trained with each.
The structure is getting more stable as the proposer learns. At step 0 three of eight programs score 0 and the rest spread between 0.023 and 0.895.
By step 3, every program is a bounded weighted sum or a mild product with the outcome term dominant, and the lowest score is 0.824.
The mean score rises from 0.44 to 0.87, since the proposer stops writing failing programs.
Moreover, one validation score has limited resolution.
The proposer writes \texttt{g1 + 0.5 * (g2 + 0.5 * (g3 + 0.5 * (g4 + 0.5)))} three times at step 3 with validation score 0.875, 0.893, and 0.879. This spread is small next to the gap between failing and working programs, which the proposer closes first.
The outer reward thus separates failing programs from working ones more sharply than it ranks working ones.
Some programs fail to execute, because of unbalanced parentheses (\textit{e.g.,} step 0 row 7) or because the generation runs to the length limit (\textit{e.g.,} step 0 row 2). Other programs execute but penalize success.
In step 0 row 1, for example, both terms are non-positive for every primitive value, and the trained policy completes no validation task.

\section{Static Proposers and Sensitivity}
\label{app:additional_experiments}

Experiments in this section train and test on GSM8K with a \texttt{Qwen2.5-1.5B} inner policy, $n=8$ rollouts (default) per outer step over seven outer steps, and report the mean over three training seeds.
``Train'' and ``Test'' denote exact-match accuracy of the final policy on the GSM8K training and test sets.
Table~\ref{tab:gsm8k_ablations} gives the final results of all four studies.

\subsection{RLAIF: Prompted GPT-4o Proposer}
\label{app:rlaif}

This baseline keeps the entire \method pipeline but replaces the trained 0.5B proposer with GPT-4o prompted in the style of Eureka~\citep{Ma2024eureka}.
At the first outer step, the prompt contains the task description, the primitive definitions, and the grammar, and asks for $n$ reward expressions.
At every later step, the prompt additionally lists the expressions proposed at the previous step and the validation score each obtained, and asks for $n$ improved expressions.
The inner-loop is unchanged, and GPT-4o receives no parameter update, so all adaptation happens in context.
Table~\ref{tab:gsm8k_ablations} shows that the prompted proposer finds a reward that beats both GRPO baselines but not \methodbase.
Its proposals do not improve across steps (Table~\ref{tab:dynamics_all}), although the prompt lists the scores of earlier proposals.


\subsection{Compute-matched Random Search}
\label{app:random_search}

Random search uses the initial proposer with frozen parameters. It has the same reward space, and training setups as \methodbase, hence the same inner-loop compute, but it never updates the proposal distribution.
Because its proposals do not improve, we select the proposal with the highest validation score over all steps and report the final policy trained with it.
Table~\ref{tab:gsm8k_ablations} shows that random search finds a reward slightly better than the outcome reward but 2.7\% below \methodbase.
Table~\ref{tab:dynamics_all} shows that its proposals do not improve across the seven steps.

\begin{table}[t]
\centering
\small
\begin{tabular}{lcc}
\toprule
\textbf{Method} & \textbf{Train} & \textbf{Test}\\
\midrule
\multicolumn{3}{c}{\textit{GRPO baselines}} \\
GRPO w/ Outcome & 89.45 & 75.82 \\
GRPO w/ Average(primitives) & 89.74 & 76.12 \\
\midrule
\multicolumn{3}{c}{\textit{Static proposers}} \\
Random search (best over steps) & 89.59 & 76.50 \\
RLAIF (prompted GPT-4o) & 89.23 & 77.18 \\
\midrule
\multicolumn{3}{c}{\textit{Removed or flipped primitives}} \\
\methodbase, flip $g_2$ & 88.52 & 77.79 \\
\methodbase, remove $g_2$ & 89.47 & 78.39 \\
\methodbase, remove $g_3$ & 89.28 & 78.92 \\
\methodbase, remove $g_4$ & 89.86 & 77.94 \\
\midrule
\multicolumn{3}{c}{\textit{Rollout size}} \\
\methodbase, $n=4$ & 89.98 & \textbf{79.23} \\
\methodbase, $n=6$ & 88.10 & 78.92 \\
\midrule
\multicolumn{3}{c}{\textit{Trained proposer, full vocabulary, $n=8$}} \\
\methodbase & \textbf{91.01} & 79.22 \\
\bottomrule
\end{tabular}
\caption{Exact match (\%) of the final policy on GSM8K in the studies of Appendices~\ref{app:rlaif}--\ref{app:number_of_rollouts}. \textbf{Bold}: best.}
\label{tab:gsm8k_ablations}
\end{table}

\subsection{Removing and Flipping Primitives}
\label{app:primitive_vocabulary}

Table~\ref{tab:gsm8k_ablations} lists the numbers behind Figure~\ref{fig:vocab}, including the GSM8K training accuracy of each final policy.

\subsection{Samples Efficiency of Proposer Evolution}
\label{app:number_of_rollouts}

We study\textit{ how many trial-and-score pairs are needed for the proposer evolution} by varying the rollout size $n$, \textit{i.e.,} number of configurations that the proposer samples at each outer step.
It sets the group over which the outer-loop update normalizes the validation scores, and it sets the cost of the search, because each configuration trains one inner policy.
We vary $n \in \{4, 6, 8\}$ in the GSM8K setting of this appendix and keep the reward space, the inner-loop budget, and the seven outer steps fixed, so the search trains 28, 42, and 56 inner policies.
Table~\ref{tab:gsm8k_ablations} reports the final policy trained with the best configuration of each search.

Final test accuracy changes by at most 0.3\% across the three sizes (78.92--79.23\%), and every size beats both GRPO baselines by 2.8--3.4\%.
Training accuracy varies more (88.10--91.01\%), with its lowest value at $n=6$.
The smallest rollout size thus matches the largest at half the inner-loop cost.
All other experiments use $n=8$.

\section{Seed Statistics for Table~\ref{tab:results_on_agent}}
\label{app:statistics_of_agent}

\begin{table}[t]
\centering
\small
\setlength{\tabcolsep}{4pt}
\begin{tabular}{cccc}
\toprule
\textbf{Outer step} & \textbf{\methodbase} & \textbf{Random search} & \textbf{RLAIF} \\
\midrule
0 & 75.13 & 76.08 & 76.88 \\
1 & 76.60 & 74.90 & 76.67 \\
2 & 77.69 & 75.97 & 76.40 \\
3 & 77.13 & 54.58 & 76.85 \\
4 & 77.29 & 76.34 & 76.45 \\
5 & 78.28 & 75.16 & 76.17 \\
6 & 78.30 & 74.83 & 76.95 \\
\bottomrule
\end{tabular}
\caption{Mean GSM8K test accuracy (\%) of the inner policies trained from the proposals at each outer step. Only the trained proposer improves over steps.}
\vspace{-8pt}
\label{tab:dynamics_all}
\end{table}

Table~\ref{tab:seed_statistics} reports the mean, standard deviation, and 95\% confidence interval ($t$-distribution) over four seeds (42, 43, 44, 45) for the methods we trained ourselves.
The intervals of \methodbase are disjoint from those of GRPO w/ Outcome at every level.
Against GRPO w/ Average(primitives) they overlap at ALFWorld L0 and L1 and at ScienceWorld L0, and are disjoint at ALFWorld L2 and ScienceWorld L1 and L2.
RLVMR, reported without seed statistics, lies outside the interval of \methodbase at ALFWorld L2 and ScienceWorld L1 and L2, and inside it at the other three levels.

\begin{table*}[t]
\centering
\footnotesize
\setlength{\tabcolsep}{4pt}
\begin{tabular}{lccc}
\toprule
\textbf{Method} & \textbf{L0} & \textbf{L1} & \textbf{L2} \\
\midrule
\multicolumn{4}{c}{\textit{ALFWorld}} \\
GRPO w/ Outcome & 76.56 $\pm$ 1.91 \ [73.52, 79.61] & 71.09 $\pm$ 2.89 \ [66.50, 75.69] & 29.69 $\pm$ 2.12 \ [26.32, 33.05] \\
GRPO w/ Average(primitives) & 88.09 $\pm$ 1.33 \ [85.96, 90.21] & 85.35 $\pm$ 1.33 \ [83.23, 87.47] & 30.47 $\pm$ 2.63 \ [26.28, 34.65] \\
\methodbase & 91.02 $\pm$ 2.07 \ [87.73, 94.30] & 89.06 $\pm$ 1.42 \ [86.80, 91.32] & 65.04 $\pm$ 2.64 \ [60.83, 69.25] \\
\methodfull & 91.80 $\pm$ 1.35 \ [89.64, 93.95] & 88.28 $\pm$ 1.78 \ [85.46, 91.11] & 76.37 $\pm$ 4.05 \ [69.92, 82.82] \\
\midrule
\multicolumn{4}{c}{\textit{ScienceWorld}} \\
GRPO w/ Outcome & 21.09 $\pm$ 3.19 \ [16.02, 26.17] & 13.67 $\pm$ 1.86 \ [10.71, 16.63] & 10.94 $\pm$ 1.10 \ [9.18, 12.70] \\
GRPO w/ Average(primitives) & 37.89 $\pm$ 4.53 \ [30.68, 45.10] & 31.25 $\pm$ 3.72 \ [25.33, 37.17] & 17.97 $\pm$ 5.45 \ [9.30, 26.64] \\
\methodbase & 47.66 $\pm$ 2.76 \ [43.26, 52.05] & 42.97 $\pm$ 2.31 \ [39.29, 46.65] & 30.08 $\pm$ 1.58 \ [27.56, 32.60] \\
\methodfull & 98.24 $\pm$ 1.61 \ [95.68, 100.00] & 95.31 $\pm$ 1.28 \ [93.28, 97.34] & 31.25 $\pm$ 2.18 \ [27.78, 34.72] \\
\bottomrule
\end{tabular}
\caption{Success rate (\%) on ALFWorld and ScienceWorld: mean $\pm$ standard deviation and 95\% confidence interval.}
\vspace{-8pt}
\label{tab:seed_statistics}
\end{table*}

\section{Computational Cost}
\label{app:cost_ana}

\paragraph{Cost breakdown.}
The inner-loop dominates the cost.
Each outer step trains $n$ inner policies, one per proposed reward, and the $n$ runs execute in parallel, each on its own set of accelerators.
We stop a proposal that fails to execute at once and set $v_i = 0$, so it costs nothing beyond waiting for the other runs.
Evaluation on $\dev$ uses vLLM and takes minutes, and so does the update of the 0.5B proposer on $n$ samples.
The wall-clock time of an outer step is therefore
\[
T_{\mathrm{step}} \approx \max_i T_{\mathrm{in}}(\phi_i) + T_{\mathrm{eval}} + T_{\mathrm{update}}.
\]

\paragraph{Budget cap.}
With some rewards the policy writes longer outputs, so inner training takes longer.
For mathematics we cap each inner run at about $1.3\times$ the FLOPs of outcome-reward training and evaluate the last checkpoint when the cap is hit. Agent runs need no such cap, because they have a fixed epoch count and a bounded action space.

\paragraph{Measured FLOPs.}
With eight inner runs of 10, 40, and 80 epochs, one outer step costs about \mbox{$7$--$9\times10^{19}$} FLOPs for mathematics, $7.3\times10^{19}$ for ALFWorld, and $1.9\times10^{20}$ for ScienceWorld.

\paragraph{Total cost.}
With $C_{\mathrm{in}}$ the cost of one inner run of the search, $C_{\mathrm{full}}$ the cost of one training run under the baseline budget, and $E_{\mathrm{out}}$ outer steps,
\begin{align*}
C_{\method\text{-base}} &\approx n \times E_{\mathrm{out}} \times C_{\mathrm{in}} + C_{\mathrm{full}},\\
C_{\method\text{-full}} &\approx n \times E_{\mathrm{out}} \times \frac{C_{\mathrm{full}}}{E_{\mathrm{out}}} + C_{\mathrm{full}},
\end{align*}
because both variants add one final training run to the search, and \methodfull divides the budget of one full run across its $E_{\mathrm{out}}$ rounds instead of repeating it.
In our runs $E_{\mathrm{out}}$ is 5--16, so with $n=8$ reward search with \methodbase costs 40--128 inner runs, and the final training of its reported policy costs one baseline run.
\methodfull runs $n \times E_{\mathrm{out}}$ short inner runs (80 runs of 10 epochs on ALFWorld) whose total compute equals $n$ full runs, and the final training of its reported policy with the same set of found rewards costs one baseline run.

\paragraph{Reducing the cost.}
With $n=4$ instead of $8$, the search cost halves at no loss on GSM8K (Appendix~\ref{app:number_of_rollouts}).
Further reductions that we have not evaluated are shorter proxy inner runs, lighter inner-loop algorithms~\citep{hu2025reinforce++, zhuang-etal-2024-efficientrag}, and searching the reward with a small inner policy before training a large one.

\section{The Outer-loop Update}
\label{app:gradient_prop}

\paragraph{The estimator.}
Let $v(\phi) = \Perf(\theta^\star(\phi);\dev)$.
The outer objective is $\mathcal{J}^{\mathrm{out}}(\psi) = \mathbb{E}_{\phi \sim \pi_\psi}[v(\phi)]$, and its gradient is
\[
\nabla_\psi \mathcal{J}^{\mathrm{out}} = \mathbb{E}_{\phi\sim\pi_\psi}\big[v(\phi)\,\nabla_\psi \log \pi_\psi(\phi\mid\ins)\big].
\]
GRPO estimates this expectation from the $n$ sampled configurations of one outer step, replaces $v(\phi_i)$ by the group-normalized advantage $(v_i - \bar v)/\sigma_v$ to reduce variance, and applies the clipped surrogate with a KL penalty to the initial proposer.
Each outer step then updates the proposer as
\[
\psi_{t+1} = \psi_t + \eta\, \widehat{\nabla_\psi \mathcal{J}^{\mathrm{out}}}(\psi_t).
\]

\paragraph{What is and is not differentiated.}
The gradient $\nabla_\psi \log \pi_\psi(\phi\mid\ins)$ is the ordinary gradient of the proposer's log-likelihood of the program it wrote.
We compute no derivative of $v$ with respect to $\phi$ or $\theta$, and none is needed, because inner training and evaluation form a black-box environment.
The same estimator trains the architecture generator in neural architecture search~\citep{zoph2017neural}.
Prompted-LLM and genetic-programming proposers do not use it, because they never update parameters from the scores.

\paragraph{Relation to meta-gradient RL.}
Meta-gradient methods~\citep{xu2018meta,zheng2018learning} compute $\partial v / \partial \phi$ by differentiating through one or more policy-update steps, which requires a differentiable, short inner update and a continuous $\phi$.
Neither holds here, because $\phi$ is a discrete program and $\theta^\star(\phi)$ is the result of hours of RL. The score-function route trades the lower variance of a pathwise gradient for applicability to this setting.

\end{document}